\documentclass[final,5p,times,twocolumn,authoryear,numafflabel]{elsarticle}

\usepackage{amsmath,amssymb}
\usepackage{graphicx}
\usepackage{etoolbox}
\makeatletter
\def\endinlineaddr{}
\newcommand\inlineaddress{\begingroup\def\par{\unskip\@ifnextchar\endinlineaddr{}{;\ }}\elsaddress\endinlineaddr\endgroup}
\patchcmd{\MaketitleBox}{\footnotesize\itshape\elsaddress\par}{\footnotesize\itshape\inlineaddress\par}{}{\PackageWarning{main}{affiliation patch failed}}
\makeatother

\usepackage{booktabs}
\usepackage{tabularx}
\usepackage{multirow}
\usepackage{xcolor}
\usepackage{hyperref}
\usepackage[normalem]{ulem}

\journal{Medical Image Analysis}

\AtBeginDocument{} %
\begin{document}

\begin{frontmatter}

\title{\huge Grounding Radiology Reports in 3D Chest CT:\\
Results of the ReXGrounding Challenge}

\author[hms]{Mohammed~Baharoon\corref{cor1}}
\author[kfh]{Mahmoud~Hussain~Alabbad}
\author[hms]{Sung~Eun~Kim}
\author[hms]{Xiaoman~Zhang}
\author[kul]{Lucas~Bijnens}

\author[uii]{Zirong~Chen}
\author[uii]{Jiayu~Wang}
\author[uii]{Xi~Ouyang}
\author[shanghaitech]{Mingyang~Yu}
\author[sim]{Weiqiu~Jin}
\author[sim]{Xiaodan~Ye}
\author[uii]{Dijia~Wu}

\author[gachon,medai]{Kwang-Hyun~Uhm}
\author[gachon]{Inhwa~Son}

\author[nki]{Rachel~van~'t~Hull}
\author[nki]{Tim~Veenboer}
\author[uva]{Cees~Snoek}
\author[nki]{Kevin~Groot~Lipman}

\author[yonsei]{Daesung~Kim}
\author[yuhs]{Junha~Im}
\author[kaist]{Youngjun~Jun}
\author[yuhs]{Sangwon~Lee}
\author[cmu]{Katelyn~Morrison}
\author[cmu]{Adam~Perer}
\author[yuhs]{Mijin~Yun}

\author[slu]{Ali~Piracha}
\author[kamc]{Reem~Alwasiah}
\author[kamc]{Talal~Yahyan}
\author[kamc]{Abdulrahman~Alamoudi}
\author[kamc]{Mashael~Alshehri}
\author[kamc]{Rayan~Alsager}
\author[nmcp]{Hannah~Hancock}
\author[mhh,breath]{Anna~M.~Hunkemöller}
\author[jlu]{Peter~Youssef}
\author[rastanura]{Mona~A.~Alhammad}
\author[rastanura]{Afnan~Aljumayan}
\author[jessa]{Brecht~Van~Berkel}
\author[jessa]{Eric~Bijnens}
\author[kfsh]{Saud~Alhussain}
\author[uzl]{Maarten~Steyvers}
\author[imelda]{Rolf~Symons}
\author[kul]{Bryan~Eelen}
\author[psmmc]{Ayed~Alaklubi}
\author[kfsh]{Maha~Buraykan~Alshammari}
\author[uoh]{Amirah~Fahad~Alshammeri}
\author[slu]{Tomas~Gilvydis}
\author[ksumc]{Ibrahim~AlTurki}
\author[ksumc]{Monerah~Almohideb}
\author[kamc]{Ali~Alburkani}
\author[kash]{Shahad~Alhomaiani}
\author[kauh]{Fatima~Jabali}

\author[dkfz,uhdmath,hidss]{Maximilian~Rokuss}
\author[dkfz,uhdmath]{Moritz~Langenberg}
\author[dkfz,uhdmath,hidss]{Yannick~Kirchhoff}
\author[dkfz,uhdmed]{Benjamin~Hamm}
\author[dkfz,uhdmath,hidss,uhdmed,helmimg,palg]{Klaus~Maier-Hein}

\author[forithmus]{Ibrahim~Ethem~Hamamci}
\author[forithmus]{Sezgin~Er}

\author[slu,ssm]{Michael~Moritz}
\author[kamc]{Hassan~R.~Alomaish}

\author[hms,yale]{Luyang~Luo\fnref{cosenior}}
\author[hms]{Pranav~Rajpurkar\fnref{cosenior}}

\affiliation[hms]{organization={Department of Biomedical Informatics, Harvard Medical School},
            city={Boston}, state={MA}, country={USA}}
\affiliation[kfh]{organization={King Fahad Hospital, Al-Ahsa Health Cluster}, city={Al Hofuf}, country={Saudi Arabia}}
\affiliation[kul]{organization={KU Leuven}, city={Leuven}, country={Belgium}}
\affiliation[uii]{organization={Shanghai United Imaging Intelligence},
            city={Shanghai}, country={China}}
\affiliation[shanghaitech]{organization={ShanghaiTech University},
            city={Shanghai}, country={China}}
\affiliation[sim]{organization={Shanghai Institute of Medical Imaging},
            city={Shanghai}, country={China}}
\affiliation[gachon]{organization={Department of Artificial Intelligence, Gachon University},
            country={South Korea}}
\affiliation[medai]{organization={MEDAI}, country={South Korea}}
\affiliation[nki]{organization={Netherlands Cancer Institute},
            city={Amsterdam}, country={The Netherlands}}
\affiliation[uva]{organization={University of Amsterdam},
            city={Amsterdam}, country={The Netherlands}}
\affiliation[yonsei]{organization={Yonsei University},
            city={Seoul}, country={South Korea}}
\affiliation[yuhs]{organization={Yonsei University College of Medicine},
            city={Seoul}, country={South Korea}}
\affiliation[kaist]{organization={Korea Advanced Institute of Science and Technology},
            city={Daejeon}, country={South Korea}}
\affiliation[cmu]{organization={Carnegie Mellon University},
            city={Pittsburgh}, state={PA}, country={USA}}
\affiliation[slu]{organization={Saint Louis University}, city={St. Louis}, state={MO}, country={USA}}
\affiliation[kamc]{organization={Medical Imaging Department, King Abdulaziz Medical City, Ministry of National Guard Health Affairs},
            city={Riyadh}, country={Saudi Arabia}}
\affiliation[nmcp]{organization={Naval Medical Center Portsmouth}, city={Portsmouth}, state={VA}, country={USA}}
\affiliation[mhh]{organization={Institute of Diagnostic and Interventional Radiology, Hannover Medical School},
            city={Hannover}, country={Germany}}
\affiliation[breath]{organization={Biomedical Research in Endstage and Obstructive Lung Disease Hannover (BREATH), German Center for Lung Research (DZL)}, city={Hannover}, country={Germany}}
\affiliation[jlu]{organization={Justus Liebig University Giessen}, city={Giessen}, country={Germany}}
\affiliation[rastanura]{organization={Ras Tanura General Hospital, Ministry of Health}, country={Saudi Arabia}}
\affiliation[jessa]{organization={Jessa Hospital}, country={Belgium}}
\affiliation[kfsh]{organization={King Fahad Specialist Hospital}, city={Buraydah}, country={Saudi Arabia}}
\affiliation[uzl]{organization={University Hospitals Leuven}, city={Leuven}, country={Belgium}}
\affiliation[imelda]{organization={Imelda Hospital}, country={Belgium}}
\affiliation[psmmc]{organization={Prince Sultan Military Medical City}, city={Riyadh}, country={Saudi Arabia}}
\affiliation[uoh]{organization={University of Hail}, city={Hail}, country={Saudi Arabia}}
\affiliation[ksumc]{organization={King Saud University Medical City}, city={Riyadh}, country={Saudi Arabia}}
\affiliation[kash]{organization={King Abdulaziz Specialist Hospital, Taif Health Cluster}, city={Taif}, country={Saudi Arabia}}
\affiliation[kauh]{organization={King Abdulaziz University Hospital}, city={Jeddah}, country={Saudi Arabia}}
\affiliation[dkfz]{organization={Division of Medical Image Computing, German Cancer Research Center (DKFZ)}, city={Heidelberg}, country={Germany}}
\affiliation[uhdmath]{organization={Faculty of Mathematics and Computer Science, Heidelberg University}, city={Heidelberg}, country={Germany}}
\affiliation[hidss]{organization={Helmholtz Information and Data Science School for Health (HIDSS4Health)}, city={Heidelberg}, country={Germany}}
\affiliation[uhdmed]{organization={Medical Faculty, Heidelberg University}, city={Heidelberg}, country={Germany}}
\affiliation[helmimg]{organization={Helmholtz Imaging}, country={Germany}}
\affiliation[palg]{organization={Pattern Analysis and Learning Group, Heidelberg University Hospital}, city={Heidelberg}, country={Germany}}
\affiliation[forithmus]{organization={Forithmus}, city={San Francisco}, state={CA}, country={USA}}
\affiliation[ssm]{organization={SSM Health}, city={St. Louis}, state={MO}, country={USA}}
\affiliation[yale]{organization={Department of Biostatistics, Yale School of Public Health}, city={New Haven}, state={CT}, country={USA}}
\cortext[cor1]{Corresponding author: \href{mailto:MohammedSalimAB@outlook.com}{\texttt{MohammedSalimAB@outlook.com}}}
\fntext[cosenior]{These authors contributed equally as senior authors.}
\begin{abstract}
\small
Segmenting each reported finding in a radiology scan would allow findings to be
measured and tracked over time and let AI systems generate reports that
are verifiable, but so far AI models have performed poorly at this
task. To advance this task, we organized the ReXGrounding Challenge at MICCAI 2026, the
first challenge dedicated to it. Given a non-contrast
chest CT scan and the set of findings from its report, participants trained
models to produce the corresponding set
of 3D segmentations. The challenge used
2,992 training and 200 validation scans from the ReXGroundingCT
dataset, and evaluated methods on a held-out test set of 300 scans with
582 findings, exhaustively annotated by
board-certified radiologists. In total, 48 teams from 15 countries made 760
submissions, and the best method reached a Dice of 0.375 and a hit rate of
0.77, matching agreement between two independent radiologist
groups (Dice 0.32, hit rate 0.76, on a 50-scan subset). On
the 100 scans of the pre-challenge benchmark, the winner reached a Dice of
0.390, against 0.285 for VoxTell, the best model before the
challenge, which five of the six best teams fine-tuned. Non-focal findings
remained far harder to segment than focal ones, instance F1 stayed below
inter-reader agreement (0.34 vs.\ 0.47), and models sometimes
segmented the right abnormality in the wrong location. Top methods thus match radiologist agreement in Dice, but non-focal findings, findings with multiple instances,
and adherence to the anatomical location described in the text remain open problems.
\end{abstract}


\end{frontmatter}
\setcounter{footnote}{0}

\section{Introduction}
\label{sec:intro}

Radiology reports describe the location of each finding only in words, such
as a ``5\,mm nodule in the posterior segment of the right upper lobe'', without
marking it on the scan, so a reader has to find it again before it can be
measured, compared with prior studies or checked. Grounding each finding, that
is, segmenting the region a sentence describes, would let findings be measured
and tracked over time and make reports, whether written by radiologists or
generated by AI, verifiable against the image
\citep{bannur2024maira,baharoon2026rexgroundingct}. This matters all the more
as AI systems for medical image interpretation advance quickly, because their
use in clinical practice depends on more than accuracy: their outputs must
also be explainable and verifiable
\citep{rajpurkar2023current,silcox2024potential,abgrall2024should}. The need is greatest in
chest CT, which is performed in large volumes (about 20 million
examinations a year in the United States alone \citep{smith2025projected})
and in which a single scan often contains many co-occurring findings, from
small nodules to diffuse emphysema.

Existing datasets, models and challenges do not support this task. Grounding
a free-text finding in 3D differs from the fixed-class segmentation
evaluated by most medical imaging benchmarks and challenges
\citep{antonelli2022medical,bilic2023liver,heller2021state,ma2024unleashing,heller2023kits21,kavur2021chaos,dorent2024lnq,luo2025segrap2023,ma2024automatic}:
a model must read a sentence that specifies the target by size, laterality, lobe
or segment and morphology, and decide which of several similar lesions it refers
to. Large chest CT datasets pair scans
with reports but lack voxel-level labels \citep{hamamci2026generalist}, and grounding datasets either link
sentences to anatomical regions rather than to findings
\citep{zhang2025development}, have fewer than 250 cases \citep{bai2024m3d}, or focus on 2D
modalities such as chest X-ray with bounding boxes instead of segmentation annotations
\citep{boecking2022making,de2025padchest}. Current text-promptable
segmentation models, trained mostly on short category-level prompts, fail on the descriptive findings
written in radiology reports
\citep{zhao2025large,zhao2025foundation,zhao2025biomedparse,du2024segvol,subramanian2026open,baharoon2025state}.
Recent challenges that pair images with radiology
reports evaluate report generation
or the localization of fixed classes of abnormalities, rather than grounding
the findings described in a report directly in the image
\citep{xu2024overview,hamamci_2025_15052708,hamamci_2026_19847782}.

\begin{figure*}[p]
\centering
\includegraphics[width=\textwidth]{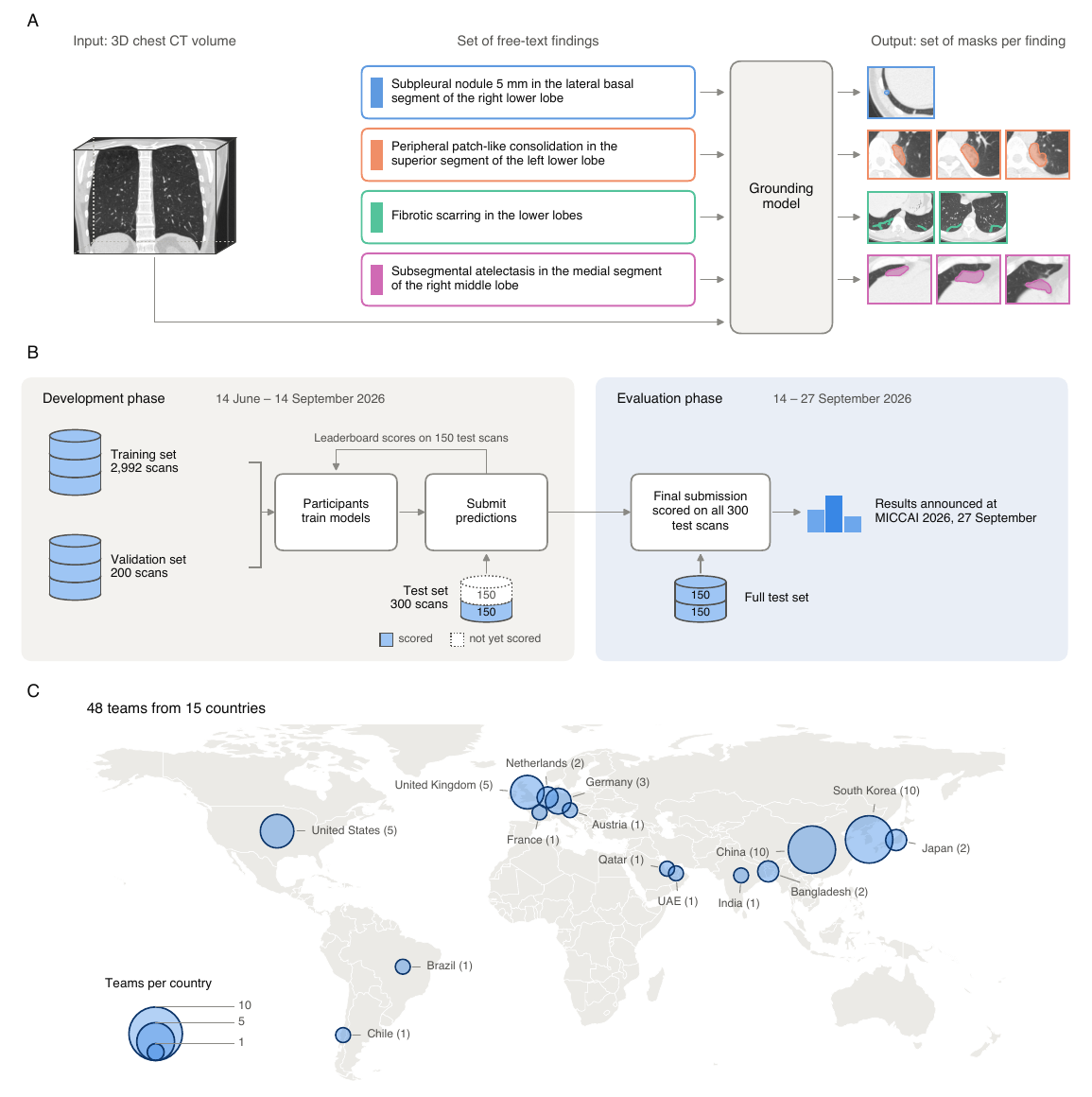}
\caption{Overview of the ReXGrounding Challenge. \textbf{A)} The task: given a
non-contrast chest CT volume and one free-text finding from its report, a model
must segment that finding in 3D. Each sentence is a separate query against the
same volume, and the shown masks are the radiologist reference annotations of a
test case (scan from CT-RATE \citep{hamamci2026generalist}). \textbf{B)} Challenge workflow. During the development phase, teams trained on the
ReXGroundingCT training set (2,992 scans with partial-instance masks)
and the validation set (200 scans with exhaustive radiologist masks), and
submitted predictions for the test set (300 scans, all visible to participants, with reference
masks withheld), and the leaderboard reported scores on a fixed half of 150 scans
(filled half; the dashed half was scored only after the deadline). After the
deadline, each team's final submission was scored on all 300 scans and the final ranking was
announced at MICCAI 2026. \textbf{C)} Participating teams by country (48 teams from 15
countries). Circle area is proportional to the number of teams. Map lines are from Natural Earth (naturalearthdata.com, version 5.1.1) and do not necessarily depict accepted national boundaries.}
\label{fig:overview}
\end{figure*}

To address this gap, we organized the ReXGrounding Challenge
(Figure~\ref{fig:overview}) at MICCAI 2026 as the first challenge on 3D
segmentation conditioned on free-text findings. The challenge is built on
ReXGroundingCT, which we previously released as the first large-scale dataset
linking findings from radiology reports to 3D segmentations in chest CT
\citep{baharoon2026rexgroundingct}. Given a non-contrast chest CT and the set of findings from its report,
participants trained models to produce the corresponding set
of 3D segmentations. Methods were evaluated on
300 held-out scans with 582 findings, exhaustively annotated by
board-certified radiologists, and 48 teams from 15 countries made
760 submissions. The challenge led to three main findings, which this paper
reports following the BIAS guideline \citep{maier2020bias}. First, the best method reached a Dice of 0.375 and a hit rate of
0.77, matching the agreement between two independent radiologist groups.
On the 100 scans of the pre-challenge benchmark it reached a Dice of
0.390 and a hit rate of 0.71, up from 0.285 and 0.62 for the best model
before the challenge \citep{rokuss2026voxtell}. Second,
the top methods performed similarly, with the top four teams within 0.015 Dice
and none statistically better than the others, and they converged on a common recipe: fine-tuning VoxTell
\citep[five of the six best teams;][]{rokuss2026voxtell}, text augmentation, ensembling, and post-processing to restrict masks to the lobe
or side named in the sentence.
Third, the top methods matched radiologists in Dice and hit rate but not at
the instance level, where the best instance F1 (0.34) stayed well below
inter-reader agreement (0.47). They also segmented focal findings far better
than non-focal ones, whose extent is also hard for radiologists to agree on,
and sometimes segmented the right abnormality in the wrong location
(Figure~\ref{fig:category-perf}). This challenge has established the first benchmark for grounding
report findings in 3D CT and substantially advanced the state of the art,
while making clear where further progress is needed. The test-set masks remain withheld behind a
public leaderboard (\url{https://rexrank.ai/ReXGroundingCT}), so that
new methods can be compared directly with the challenge entries.

\section{Challenge design}
\label{sec:methods}

The ReXGrounding Challenge was held as part of MICCAI 2026 and hosted on the
ReXrank leaderboard platform \citep{zhang2024rexrank}
(\url{https://rexrank.ai/ReXGroundingCT}). Its design was registered in advance
with MICCAI \citep{baharoon2026design}. The organizers, timeline, submission
procedure and policies are described in \ref{app:organization}.

\begin{figure*}[t]
\centering
\includegraphics[width=\textwidth]{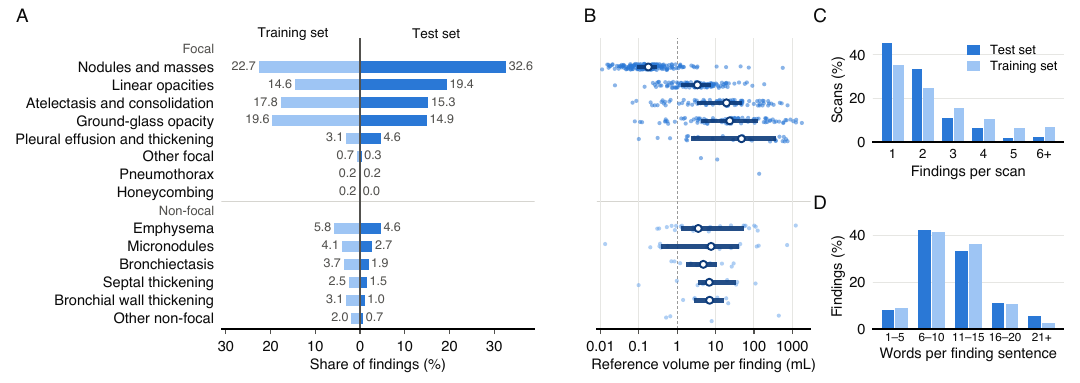}
\caption{Challenge data. A) Finding categories as a share of all findings in
the training set (left, 7,687 findings in 2,992 scans) and the
test set (right, 582 findings in 300 scans), grouped into focal and
non-focal categories. Numbers give the share in percent. B) Reference volume of every test-set finding by category (dots). For categories
with at least five findings, the bar spans the interquartile range and the open
circle marks the median. The dashed line marks 1\,mL. C) Number of findings per
scan. D) Length of the finding sentences. In C and D, dark bars show the test
set and light bars the training set.}
\label{fig:categories}
\end{figure*}

\subsection{Task definition}
\label{sec:mission}

\paragraph{Task}
Given a non-contrast chest CT volume and the set of free-text findings from
its report, produce the corresponding set of 3D binary segmentations, one per
finding, each covering all regions that finding describes. When a finding has
several instances, such as multiple nodules, all of them are segmented
together in its mask. The reference masks label each instance separately for
the instance-level metrics (Section~\ref{sec:metrics}).

\paragraph{Assessment aim}
The challenge assessed how well each finding described in the text was
localized and delineated in 3D. The primary metric was Dice, the overlap between
the predicted and reference masks of each finding, which reflects both whether
the finding was found and how well its extent was captured. The secondary
metrics were the hit rate, which measures whether a model localized each finding
at all regardless of how precisely it outlined it, and instance F1, which
measures whether it found each lesion of findings made up of several lesions.
All metrics are defined in Section~\ref{sec:metrics}.

\subsection{Challenge data}
\label{sec:data}

\paragraph{Data source}
Scans and reports come from CT-RATE, a public dataset of
de-identified non-contrast chest CT scans and their radiology reports acquired
during routine care at Istanbul Medipol University Mega Hospital, Turkey
\citep{hamamci2026generalist}. The original Turkish reports had been
machine-translated into English and corrected by bilingual medical students in
CT-RATE. For ReXGroundingCT, GPT-4 rewrote each report into standard US
radiology terminology and split it into individual findings, and GPT-4o-mini
assigned each finding to a category by type of abnormality
\citep{baharoon2026rexgroundingct}. The lung and pleural findings used in the
challenge fall into 14 such categories, such as nodule, emphysema or pleural
effusion, grouped into focal and non-focal findings
(Figure~\ref{fig:categories}). A board-certified radiologist confirmed the
accuracy of each step on random samples \citep{baharoon2026rexgroundingct}. The finding sentences used in the challenge
are the outputs of this pipeline and were not edited further.

\paragraph{Data splits}
A case is one CT volume with its set of findings.
The training set contains 2,992 scans (7,687 findings) with partial-instance
annotations (up to three instances per finding). The validation set
(200 scans, 381 findings) and the test set (300 scans, 582 findings) were
exhaustively annotated by board-certified radiologists. During the challenge,
the leaderboard showed results on a fixed random half of the test set
(150 scans, 302 findings, the leaderboard half), and the final ranking used all
300 scans.
Figure~\ref{fig:categories} summarizes the composition of the training and
test sets. The test set has a slightly higher share of focal findings than the
training set (88\% vs.\ 79\%), mainly because nodules and masses are more common (33\%
vs.\ 23\%).
The training set is identical to that of the original ReXGroundingCT release
\citep{baharoon2026rexgroundingct}. The challenge
validation set contains the 50 original validation scans plus 150
new scans, and the challenge test set contains
the 100 original test scans plus 200 new scans. The new scans were annotated
with the same exhaustive protocol as the original ones, by an additional group of
annotators.

\paragraph{Annotation}
All masks were drawn on a cloud-based, HIPAA-compliant annotation platform
with brush, polygon and region-growing tools, adjustable window settings and
free navigation across slices \citep{baharoon2026rexgroundingct}. Training scans followed a
partial-instance protocol, in which at most three representative instances of
each finding were segmented. 1,400 training scans were annotated by
professional annotators and accepted, corrected or returned by board-certified
radiologists, and 1,592 by trained medical students supervised by
radiologists until they could annotate independently. Validation and test scans
followed an exhaustive protocol, in which every visible instance was segmented.
The original 150 validation and test scans were annotated by four
board-certified radiologists, and the 350 scans added for the challenge by
an additional group of annotators under the same protocol. Findings that could
not be localized on the scan, were too diffuse to outline
(for example, mosaic attenuation throughout the lungs), or described
normal anatomy were excluded.

\paragraph{Pre-processing}
No pre-processing was applied. Participants downloaded the original scans
from CT-RATE as NIfTI volumes, and each reference mask has the same array
dimensions as its scan, with one channel per finding. Participants chose their
own resampling, cropping and intensity normalization.

\paragraph{Sources of error}
The main source of error in the reference standard is variability between
annotators. In an inter-rater study on 50 validation and test scans
(96 findings) annotated twice by non-overlapping groups of radiologists,
agreement reached a Dice of 0.32, a hit rate of 0.76 and an instance
F1 of 0.47 \citep{baharoon2026rexgroundingct}.
For focal findings, agreement reached a Dice of 0.34 and a hit rate of
0.84. For non-focal findings it was lower (Dice 0.15, hit rate
0.41), although with only 17 such findings in the study this
estimate is imprecise. The lower agreement is consistent with the extent of a
diffuse abnormality being inherently ambiguous. Errors can also come
from the reports, through translation from Turkish and the LLM rewriting and
extraction steps.
One participating team (Team 4) reported correcting more than 100 training
annotations, including masks on the wrong side and findings absent from the
source report, which they used in their own solution
(Section~\ref{sec:top-methods}). Finally, the training and
evaluation sets were annotated under different protocols. Training masks
cover at most three instances of each finding, so further instances are
unlabelled and appear as background, whereas the evaluation masks include
every instance. A model trained on the partial masks can therefore learn to
leave some instances unsegmented and is penalized for missing them at test time.

\paragraph{Cohort}\label{sec:cohorts}
The target cohort is adult patients undergoing non-contrast chest CT in routine
clinical practice, and the challenge cohort was drawn from this population. Test-set
patients had a median age of 41 years (interquartile
range 31--51, range 18--89) and 42\% were female, similar to
the training set (median 40 years, 44\% female) and the validation
set (median 37 years, 44\% female). Test scans were acquired between
2016 and 2022 on six scanner models from Philips, Siemens and
Philips--Neusoft, with a median slice spacing of 1.0~mm (range
0.75--2.0) and a median in-plane spacing of 0.71~mm (range
0.39--0.98). Each scan covers the full chest, and the
segmentation targets are the lung and pleural findings described in its
report, grouped into 14 categories
spanning focal and non-focal patterns (Figure~\ref{fig:categories}). No clinical information
beyond age and sex was provided.

\subsection{Challenge rules}
\label{sec:organization}

All predictions on the test set had to be fully automatic, without manual
editing or case-specific tuning. Participants could use any public or private
training data, pretrained models, and additional annotations, provided these
were described in their method submission. Training on test cases, including
self-supervised training, was not allowed. The challenge had two tracks, which differed in how the finding text could
be processed. In the Main track (free-text grounding), methods had to take the
finding text exactly as provided, optionally supplemented by its category label.
The Overall track also admitted methods that first converted the text into a
structured representation, for example by extracting location, morphology or
size with a language model or parser, and every eligible Main-track submission
was also ranked in the Overall track. In both tracks, each mask had to be
derived from the text of its finding: methods that segmented only by category,
ignoring the rest of the sentence, were not allowed. A method counted as text-conditioned if the mask it produced for a finding
could depend on the wording of the sentence beyond its category, so that two
findings of the same category with different descriptions, for example in
different lobes, could receive different masks. A fixed-class model that maps
each category label to a single mask was therefore not allowed, even though
the category labels were provided. Methods could use anatomical segmentations as
additional inputs and post-process masks, for example by restricting them to the
anatomical region named in the finding. All six best-performing
teams submitted to the Main track, so the two tracks produced the same winners,
and we report a single ranking.

\subsection{Assessment method}
\label{sec:assessment}

\paragraph{Metrics}\label{sec:metrics}
The primary metric was Dice, computed for each finding over its whole mask,
averaged over the findings in each case, and then averaged over cases. Before
any metric was computed, predicted connected components smaller than 10
voxels were removed and at most the 50 largest components per finding were
kept. An empty prediction for a finding scored a Dice
of zero. We defined the hit rate as the proportion of findings
with a Dice of at least 0.1. For the instance-level metrics, the filtered
prediction was split into connected components, which were matched one to one
with the annotated reference instances in order of decreasing Dice, and a match counted as a true
positive when its Dice was at least 0.2. Instance precision, recall and F1
follow from these matches.

Dice over the whole mask was used as the single primary metric because it
captures both where a finding is and its extent, such as its volume and shape,
which matter for measuring findings and tracking them over time. Following Metrics Reloaded
\citep{maier2024metrics}, Dice was complemented by the detection metrics hit rate
and instance F1.
Averaging per case before averaging over cases gives every scan equal weight,
so scans with many findings do not dominate. Dice is, however, a poor fit for
diffuse findings, whose extent is inherently ambiguous
(Section~\ref{sec:data}). Results for these findings should therefore be read
alongside the hit rate, and we return to this in the Discussion.

\paragraph{Ranking}
For each team we ranked the submission it declared as final or, if none was
declared, its best-scoring submission on the leaderboard half of the test set.
8 of the 48 leaderboard entries (teams) declared a final
submission, and the remaining 40 defaulted to their best leaderboard-half
submission. Teams were ranked by the primary metric computed on the full
300-scan test set, following an aggregate-then-rank scheme: each
team's Dice was first averaged over all findings and cases, and teams were
then ordered by this mean. Missing cases could not occur in the ranking, because a submission was scored
only if, for each of the 300 test scans, it contained one mask per
finding with the same dimensions as the scan.
No two teams had the same mean Dice, so there were no ties. We chose aggregate-then-rank because it reflects how much better one method is than
another, whereas ranking teams on each scan and averaging their ranks
(rank-then-aggregate) reflects only how often it is better. As a sensitivity analysis, we also ranked
the teams with rank-then-aggregate (Section~\ref{sec:stability}).

\paragraph{Statistical analysis}
We computed 95\% confidence intervals for each team's mean Dice, hit rate and
instance F1 from 1,000 bootstrap resamples of the 300 test cases. They are given in
Table~\ref{tab:leaderboard} and Table~\ref{tab:leaderboard-full}. Following
\citet{maier2018rankings} and \citet{wiesenfarth2021methods}, ranking
variability was assessed by re-ranking all teams in each replicate and
reporting the distribution of each team's rank and the range covering 95\% of
its bootstrap ranks (Figure~\ref{fig:ranking}A), and agreement between
bootstrap and observed rankings was summarized with Kendall's $\tau$. For each pair among the ten best teams, we tested whether the higher-ranked
team scored significantly higher with a one-sided paired Wilcoxon signed-rank
test on per-case Dice, with Holm correction for the 45 comparisons
(Figure~\ref{fig:ranking}B), and report the proportion of replicates in which
it scored higher. All
analyses used Python with NumPy and SciPy.

\subsection{Prompt sensitivity analysis}
\label{sec:prompt-exp}
To test how the location, size and other details given in a finding sentence
affect the predicted mask, the final methods of the
best teams were run on edited versions of the test
sentences, with two kinds of edit, removing information and introducing wrong
information. Information was removed by
deleting the stated size, the lobe or segment, or the side, or by keeping only
the name of the finding. Wrong information was introduced by swapping left and
right, naming a different lobe in the same lung, or changing a small stated size
to 30\,mm and a large one to 3\,mm. Findings in both lungs were also
prompted once for each side. Size and side edits were made with rules, and the
other edits were made by a large language model (Claude Opus 5.5, Anthropic).
Each edit was compared with the original sentence on the findings it changed,
using the paired difference in Dice and a two-sided Wilcoxon signed-rank test
with Holm correction over the edits of each team. For
the location edits we also measured the share of the predicted mask that lay on
each side of the body. The effect of each edit on Dice and on the location of the
mask is reported in Section~\ref{sec:further}, and Figure~\ref{fig:prompt-examples} shows example
masks predicted for original and edited sentences.

\section{Results}
\label{sec:results}

Throughout this paper, teams are numbered by their final rank, so Team 1 is the
winning team and Team 10 the tenth-placed team (Table~\ref{tab:leaderboard}).

\subsection{Participation}
\label{sec:participation}
\begin{figure}[!tp]
\centering
\includegraphics[width=\columnwidth]{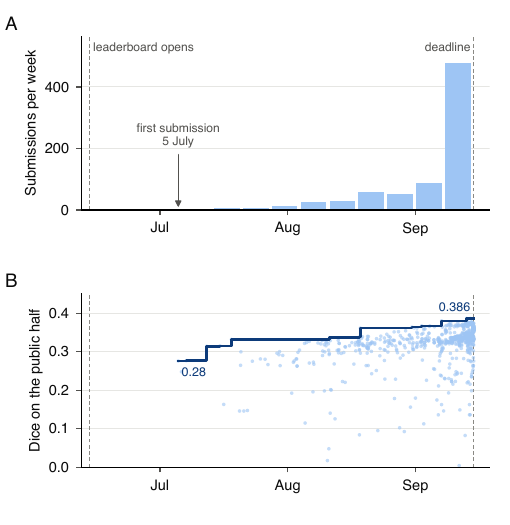}
\caption{Submissions over the development phase. A) Scored submissions per
week, in seven-day windows ending at the deadline. B) Dice on the leaderboard half of the test set for every scored
submission (dots) and the best score reached so far (line). Dashed lines mark
the opening of the leaderboard and the submission deadline.}
\label{fig:timeline}
\end{figure}

Submissions came from 53 team registrations. Some groups had registered
more than once, so we merged registrations with overlapping team members
(or, in one case, the same team name and institution) into a single team,
keeping that team's
best-scoring final submission. After merging, 48 teams from 15
countries had made 760 submissions in total (median 4 per team).
China (including Hong Kong) and South Korea were the most represented
countries, with 10 teams each (Figure~\ref{fig:overview}C).
We classified each team by the affiliations it listed at registration for
the team and its members. 35 of the 48 teams listed only
universities, research institutes or hospitals, 6 listed only
companies, 4 listed both a company and a university or hospital, and 3 did not specify an
affiliation. Industry involvement was concentrated at the
top of the ranking. Five of the six best-performing teams had a company among
their affiliations, with Team~3 as the only exception. Submissions were concentrated at the end of the development
phase: the first submission to appear on the leaderboard arrived on 5 July 2026, three weeks after the
leaderboard opened, and 74\% of all submissions were made in the final
two weeks (Figure~\ref{fig:timeline}A). The best score on the leaderboard half rose
from 0.28 for the first submission to 0.386 on 13 September,
the day before the deadline (Figure~\ref{fig:timeline}B).

\begin{figure}[!tp]
\centering
\includegraphics[width=\columnwidth]{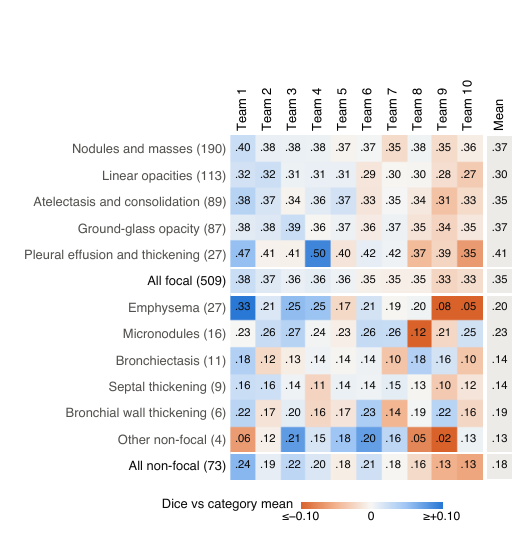}
\caption{Mean Dice per finding of the ten best teams by finding category on the
full test set. Numbers in cells give the Dice, and colour shows how far each team is
above (blue) or below (orange) the mean of the ten teams for that category. The last column gives that mean.
Numbers in brackets give the number of test findings. Pneumothorax (1
finding) and other focal findings (2) are included in the focal total but
not shown.}
\label{fig:category-perf}
\end{figure}

\begin{table}[t]
\centering
\caption{Final ranking on the full 300-scan test set (top 10). Teams are numbered by their final rank. CI,
95\% bootstrap confidence interval over the test scans (1,000 replicates,
Section~\ref{sec:stability}). The full leaderboard is in \ref{app:leaderboard}.}
\label{tab:leaderboard}
\renewcommand{\baselinestretch}{1}\footnotesize
\setlength{\tabcolsep}{3pt}
\renewcommand{\arraystretch}{1.2}
\begin{tabular}{lccc}
\toprule
Team & Dice {\scriptsize[95\% CI]} & Hit rate {\scriptsize[95\% CI]} & Instance F1 {\scriptsize[95\% CI]} \\
\midrule
Team 1 & 0.3755 {\scriptsize[0.352--0.399]} & 0.7680 {\scriptsize[0.732--0.802]} & 0.3410 {\scriptsize[0.312--0.369]} \\
Team 2 & 0.3654 {\scriptsize[0.344--0.387]} & 0.7629 {\scriptsize[0.730--0.797]} & 0.3428 {\scriptsize[0.316--0.371]} \\
Team 3 & 0.3621 {\scriptsize[0.342--0.383]} & 0.7646 {\scriptsize[0.737--0.795]} & 0.3336 {\scriptsize[0.306--0.364]} \\
Team 4 & 0.3605 {\scriptsize[0.339--0.382]} & 0.7766 {\scriptsize[0.744--0.809]} & 0.3116 {\scriptsize[0.285--0.341]} \\
\midrule
Team 5 & 0.3526 {\scriptsize[0.330--0.375]} & 0.7457 {\scriptsize[0.713--0.781]} & 0.1899 {\scriptsize[0.166--0.216]} \\
Team 6 & 0.3519 {\scriptsize[0.332--0.373]} & 0.7646 {\scriptsize[0.732--0.799]} & 0.2435 {\scriptsize[0.217--0.271]} \\
Team 7 & 0.3432 {\scriptsize[0.322--0.366]} & 0.7526 {\scriptsize[0.721--0.785]} & 0.2611 {\scriptsize[0.238--0.288]} \\
Team 8 & 0.3393 {\scriptsize[0.317--0.362]} & 0.7423 {\scriptsize[0.708--0.777]} & 0.2463 {\scriptsize[0.220--0.275]} \\
Team 9 & 0.3329 {\scriptsize[0.313--0.354]} & 0.7165 {\scriptsize[0.683--0.750]} & 0.2481 {\scriptsize[0.222--0.278]} \\
Team 10 & 0.3329 {\scriptsize[0.311--0.355]} & 0.6890 {\scriptsize[0.654--0.726]} & 0.2985 {\scriptsize[0.269--0.328]} \\
\bottomrule
\end{tabular}
\end{table}

\subsection{Overview of the top-performing methods}
\label{sec:top-methods}
All three winning teams followed the same recipe. A frozen Qwen3-Embedding
text encoder \citep{zhang2025qwen3} turns the finding sentence into a single
embedding vector, and a 3D segmentation network takes the CT volume together
with this vector and outputs the mask of that finding. The teams differed in
five places. The first difference is the image encoder. Team 1 used the chest
CT foundation model CIRCLE\footnote{\url{https://github.com/uii-open-source/CIRCLE}}.
Team 2 used a residual-encoder nnU-Net trained from scratch. Team 3 used the
VoxTell encoder together with a second model, TAP-CT \citep{veenboer2025tap}.
The second difference is how the text embedding is combined with the image
features. Teams 1 and 3 kept the VoxTell decoder \citep{rokuss2026voxtell}, in
which the text embedding acts as the query of a small transformer that
cross-attends to the image features at the bottleneck. Its output is projected
to every decoder stage, where it is multiplied channel-wise with the image
features and the product is concatenated back onto them, so the text steers
the mask at every resolution. Team 2, and Team 3's TAP-CT branch, used FiLM
\citep{perez2018film} instead, in which small networks map the embedding to a
per-channel scale and shift that are applied to the image features, with no
cross-attention between text and image. The
third difference is the training data. Team 1 added 950 in-house annotated
scans, Team 2 used only the challenge data, and Team 3 added 8,000 CT-RATE
sentences with pseudo-masks for two of its ensemble members. The fourth
difference is ensembling. Team 1 averaged checkpoints of a single model, Team 2
averaged 13 models, and Team 3 combined 17 (twelve VoxTell and five TAP-CT
models). The fifth difference is post-processing. Teams 1 and 2 only
thresholded the probabilities, whereas Team 3 restricted each mask to the lobe
named in the sentence, chose a threshold per finding and removed components
predicted to lower Dice. More detailed descriptions of each method are given in \ref{app:methods}.

\paragraph{Team 1} The winning entry paired the image
encoder of CIRCLE, a chest CT foundation model pretrained on 403,500
non-contrast scans with paired reports from 288,740 patients through
image--report contrastive learning and supervised classification of 37
abnormalities, with the text-prompted segmentation framework of VoxTell
\citep{rokuss2026voxtell}. The image encoder and the VoxTell
decoder were fine-tuned on the challenge training set, and the text encoder was
kept frozen. Scans were resampled to 0.65 $\times$ 0.65 $\times$ 1.15~mm
and trained on 384 $\times$ 320 $\times$ 224 voxel inputs with a
Dice--cross-entropy loss. On a 50-case validation subset, pretraining raised
mean Dice from 0.277 (same architecture from scratch) to 0.327, and
adding in-house data (800 pulmonary-nodule segmentations and 150
cases of less common thoracic disease) raised it further to 0.356. The
final submission averaged checkpoints of a single model from several epochs.

\paragraph{Team 2} The second-placed team used no external data
and a simple architecture they called Grounder: a residual-encoder
3D nnU-Net \citep{isensee2021nnu,isensee2024nnu}, conditioned on the finding
text through feature-wise linear modulation \citep{perez2018film}. The sentence was embedded with Qwen3-Embedding-4B
\citep{zhang2025qwen3}, and a two-layer MLP mapped the embedding to
scale and shift parameters applied at every decoder stage. Two choices drove
most of the gain. First, a large input patch of 448 $\times$ 448 $\times$
256 voxels gave the network enough anatomical context to resolve location
phrases. Second, in-distribution text augmentation replaced each finding
sentence, with some probability, by a real sentence from another training case
that described the same finding type at the same location. The final
submission averaged 13 Grounder models trained with different patch
sizes, which the team found to be complementary. Adding lung-lobe masks from
TotalSegmentator as input channels, an explicit finding-category embedding, and
negative text examples did not help.

\paragraph{Team 3} The third-placed team combined two model
families. The main branch was an ensemble of 12 VoxTell models, in which
the text embedding (Qwen3-Embedding-4B) enters an nnU-Net decoder as a query
through cross-attention, trained on 192$^3$ patches. The second branch
was an ensemble of five models that used TAP-CT \citep{veenboer2025tap}, a 3D
vision transformer pretrained in the style of DINOv2 \citep{oquab2023dinov2},
as the image encoder, with a convolutional skip decoder and FiLM conditioning on
Qwen3-Embedding-0.6B. This branch was weaker alone but added diversity, and the
two branches were fused with weights of 0.8 and 0.2. All members were trained on the training and validation sets
with synonym-based text augmentation (31 synonym groups), but their recipes
otherwise varied. Two of the 12 VoxTell members added 8,000 CT-RATE
sentence--mask pairs with pseudo-masks of pleural effusions and nodules from
TotalSegmentator \citep{wasserthal2023totalsegmentator}, one added lobe
segmentation as an auxiliary task, and five oversampled the validation set
five-fold late in training. Post-processing had four steps. Each mask was
restricted to the lobe named in the sentence (mapped offline with an LLM and
segmented with TotalSegmentator), or to the named lung when no lobe was given.
A learned model then set a separate threshold for each finding, a small 3D CNN
removed components predicted to lower Dice, and voxels claimed by several
findings went to the most confident one. The team found the lobe restriction,
the per-finding thresholds and TAP-CT as a second architecture to be the most
helpful.

\paragraph{Team 4} The fourth-placed team also fine-tuned
VoxTell, on 288 $\times$ 288 $\times$ 192 crops at 0.75 $\times$ 0.75
$\times$ 1.25~mm, and reported that fine-tuning beat training the same
architecture from scratch by more than 0.05 Dice. Three further steps
improved performance. They audited the training annotations and corrected more
than 100 cases, including masks on the wrong side, drawing artifacts,
and findings that the source report did not contain. They extracted
62,705 finding sentences from 18,652 CT-RATE reports with MedGemma
\citep{sellergren2026open} and used these scans for mean-teacher
semi-supervised training, together with external segmentation datasets and a
presence loss on pooled voxel logits. Finally, each model was fine-tuned for
a further 500 steps on the validation set alone, whose exhaustive
annotation protocol is closest to the test set. As a post-processing step,
predictions for sentences naming a side were trimmed to that lung. The team reported no gain from a Tversky loss
intended to tolerate partial annotations, and no improvement on
bronchiectasis and micronodules, which they attributed to annotation
variability. Their entry had the highest hit rate of all teams
(0.777).

\paragraph{Teams 5 and 6} The next two teams were within 0.01 Dice of
fourth place. Team 5 combined two VoxTell-based 3D models with a model that
segments native-resolution axial slices. Both 3D models also received a coarse
map of where the finding is likely to lie, from a vision--language model the
team pretrained on CT-RATE. Because the training masks cover at most three
instances of a finding, the team excluded uncertain lung voxels from the loss
rather than treating them as background, and at inference each mask was
restricted to the side or lobe named in the sentence, keeping only the most
probable component when the sentence described a single object. Team 6 extended
VoxTell with several queries formed from the individual words of the sentence,
weighted by the whole-sentence embedding, and trained with a loss that
penalises predictions inside a finding of another category in the same scan. They used
specialist models for selected categories such as nodules, and removed
predictions outside the named lobe.

\subsection{Final ranking}
\label{sec:ranking}
Table~\ref{tab:leaderboard} gives the final ranking of the ten best teams, and
Table~\ref{tab:leaderboard-full} that of all 48. Team 1 won with a mean Dice of
0.3755, followed by Team 2 (0.3654) and Team 3 (0.3621). The secondary metrics
ordered the top teams differently. Team 4 had the highest hit rate (0.7766) and
Team 2 the highest instance F1 (0.3428).
Dice across all teams ranged from 0.146 to 0.375 (median
0.280). The top four teams lay within 0.015 Dice, and
3rd and 4th place were separated by 0.0016. The overall ranking also masks large
differences between finding categories, which are shown in
Figure~\ref{fig:category-perf}. Performance did not differ significantly by
patient sex or age group for any of the ten best teams (\ref{app:sexage}).

\paragraph{Comparison with models before the challenge} Before the challenge,
the public ReXGroundingCT leaderboard evaluated models on a 100-scan
subset of the challenge test set \citep{baharoon2026rexgroundingct}. General
text-promptable segmentation models, which are trained on names of anatomical
structures and lesion classes rather than on descriptive findings, failed out
of the box \citep{baharoon2025state}. SAT \citep{zhao2025large}, SegVol \citep{du2024segvol} and
BiomedParse-V \citep{zhao2025biomedparse} all reached a Dice close to 0. Fine-tuning SAT on the ReXGroundingCT training set raised its Dice
to 0.205 (hit rate 0.473), and the best entry at the time, VoxTell
\citep{rokuss2026voxtell}, reached 0.285 (hit rate 0.615). On the
same 100 scans, the winning team reached a Dice of 0.390 (hit rate
0.712), and 23 of the 48 teams scored above VoxTell.

\begin{figure*}[!tp]
\centering
\includegraphics[width=\textwidth]{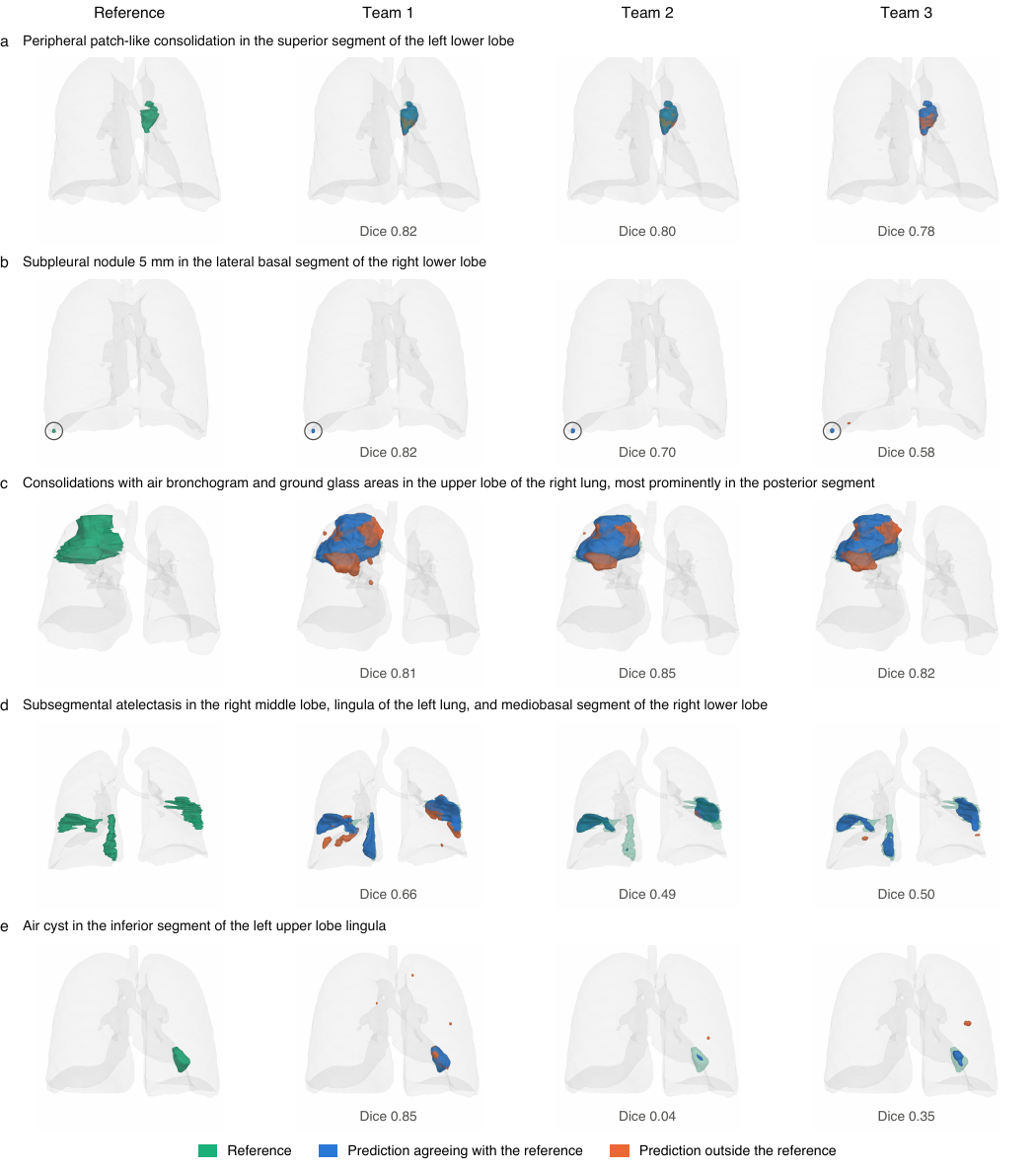}
\caption{Examples of successful grounding by the three winning teams. Masks are rendered
in 3D inside the lungs and viewed from the front (patient's right on the left).
The reference is shown in green, solid in the first column and translucent in
the others. Predicted surfaces are blue where they lie within 3\,mm of the
reference and orange elsewhere. Lesions smaller than 1\,mL are circled.
Dice is computed over the whole 3D mask of the finding.}
\label{fig:qual-success}
\end{figure*}

\subsection{Ranking stability}
\label{sec:stability}
The ranking on the leaderboard half of the test set closely matched the final
ranking on all 300 scans. The first two places were unchanged,
and places three to eight changed only by swaps between neighbouring pairs.
The ranking was stable overall, and bootstrap rankings agreed closely with
the observed one (Figure~\ref{fig:ranking}, median Kendall's
$\tau = 0.95$). At the top, however,
the teams performed too similarly to be separated reliably. Team 1 ranked
first in 89\% of 1,000 bootstrap replicates and within the top three in
99\%, but the
95\% confidence intervals of the top four teams overlapped (Table~\ref{tab:leaderboard}), and none of the differences among them was
significant after correction for the 45 pairwise comparisons. Team 1 scored higher than Team 2 in
91\% of replicates, whereas Team 3 beat Team 4 in only 57\%. The
winner was significantly better than the fifth- and sixth-placed teams
(adjusted $p = 0.017$ and 0.009).
Ranking with rank-then-aggregate instead of the official aggregate-then-rank
scheme (Section~\ref{sec:metrics}) gave a similar ranking overall (Kendall's
$\tau = 0.96$) and the same four teams at the top, but in the order
Teams 2, 3, 1 and 4. Team 1 had the highest mean Dice, whereas Team 2 had
the best average rank across scans.

\subsection{Performance by finding type}
\label{sec:by-category}

Focal findings were segmented far better than non-focal findings by all
top teams: across the six best teams, Dice on focal findings (n=509) ranged from
0.348 to 0.378 with hit rates of 0.78--0.82, whereas Dice on non-focal
findings (n=73) ranged from 0.176 to 0.243 with hit rates of
0.47--0.53. Focal performance was therefore tightly clustered, and the winner's
margin came largely from non-focal findings (0.243 vs.\ at most 0.216 for the
other top-six teams).
Figure~\ref{fig:category-perf} breaks these results down by category. Among
focal findings, pleural effusion and thickening was segmented best (Dice
0.35--0.50) and linear opacities worst (0.27--0.32). Non-focal categories varied far
more between teams than focal ones. The winner's non-focal lead came mainly from emphysema,
where Team 1 reached 0.33 and no other top-ten team exceeded
0.25. Bronchiectasis and septal thickening were hard for every team (Dice at most
0.18), and the smallest non-focal
categories, with 4 to 9 findings each, were too small to rank teams
reliably.

\subsection{Qualitative comparison}
\label{sec:qualitative}
Figure~\ref{fig:qual-success} shows the final submissions of the three
winning teams on five findings that were grounded well, chosen to cover
different finding types. A
consolidation in the left lower lobe and a 5\,mm subpleural nodule (Figure~\ref{fig:qual-success}a and b) were localized by all three teams (Dice 0.58--0.82), with differences mainly in
boundary detail. The teams' solutions also followed longer, more detailed sentences. For
consolidations with air bronchograms and ground-glass areas in the right upper
lobe, most prominent in the posterior segment (Figure~\ref{fig:qual-success}c), all three reached a Dice above
0.80. For subsegmental atelectasis at three named sites in both lungs (Figure~\ref{fig:qual-success}d),
Teams 1 and 3 placed a mask at every named site and Team 2 at two of them (Dice
0.49--0.66). Other findings separated the teams. Team 1 segmented an
air cyst in the lingula (Figure~\ref{fig:qual-success}e) almost completely (Dice 0.85), whereas Teams 2 and
3 marked only part of it.

\begin{figure*}[!tp]
\centering
\includegraphics[width=\textwidth]{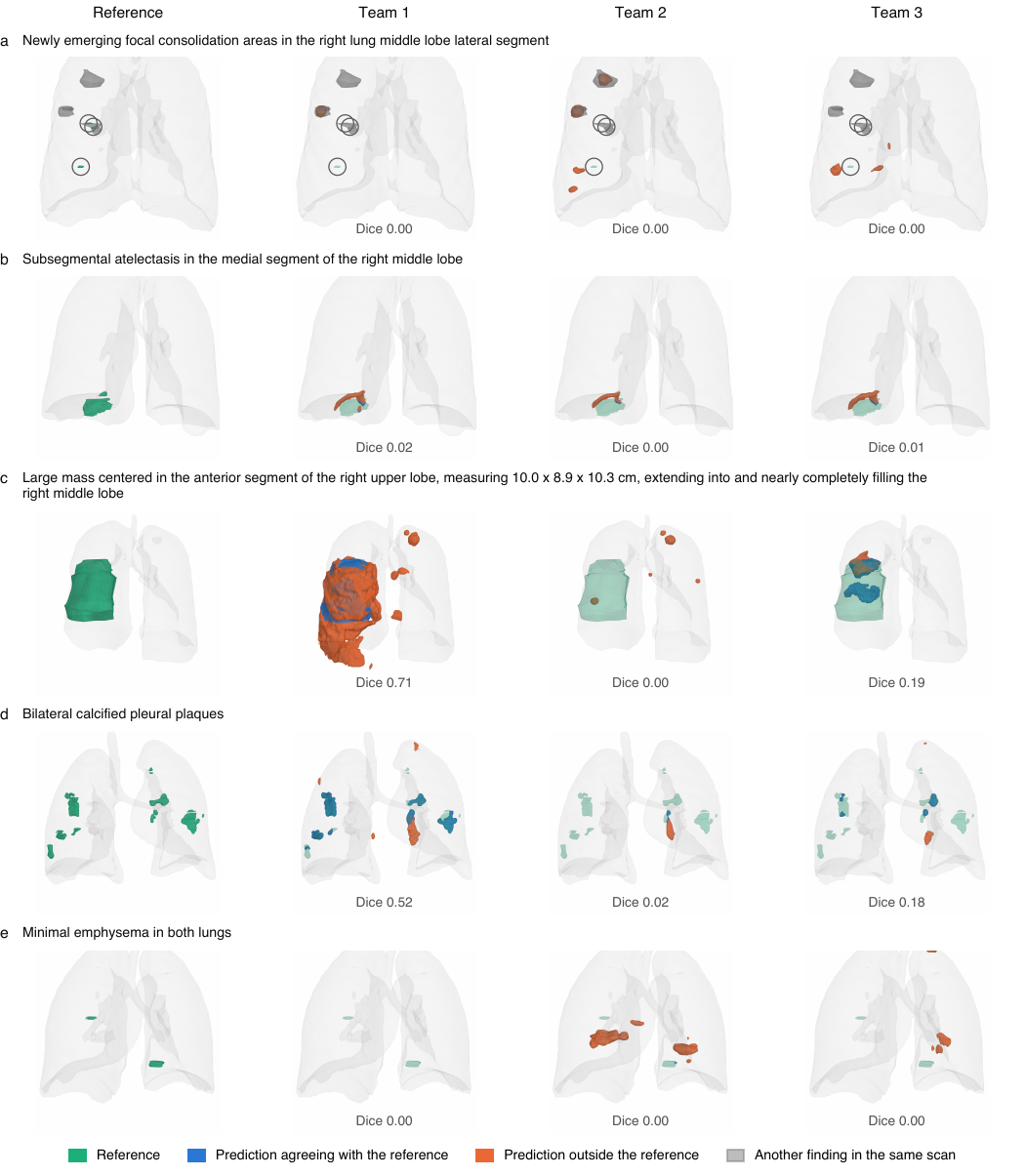}
\caption{Examples of grounding failures by the three winning teams, shown as in
Figure~\ref{fig:qual-success}. In a, other reference findings of the
same scan are shown in grey: two of the teams segmented a consolidation that the
report describes as a separate finding in the right upper lobe.}
\label{fig:qual-failure}
\end{figure*}

Figure~\ref{fig:qual-failure} shows five findings that were grounded poorly,
each illustrating a different problem. For focal consolidation described
in the lateral segment of the right middle lobe (Figure~\ref{fig:qual-failure}a), Teams 1 and 2 instead
segmented a consolidation in the right upper lobe that the same report lists as
a separate finding (Dice 0.19 and 0.46 against that finding), so the
models recognized the abnormality but not the location named in the sentence.
For subsegmental atelectasis in the medial segment of the right middle lobe
(Figure~\ref{fig:qual-failure}b), all
three teams segmented a thin neighbouring structure rather than the atelectatic
segment (Dice at most 0.02). Team 1 was the only team to
capture a 10\,cm mass in the right upper lobe (Figure~\ref{fig:qual-failure}c, Dice 0.71), although its
mask extended beyond the reference, while Team 2 missed it entirely. Calcified pleural plaques (Figure~\ref{fig:qual-failure}d) were found only in
part by Team 1 (Dice 0.52) and largely missed by the other two teams.
For minimal emphysema, a non-focal finding (Figure~\ref{fig:qual-failure}e), Team 1 predicted nothing and the
other two teams marked regions away from the annotated foci.

\begin{figure*}[!tp]
\centering
\includegraphics[width=\textwidth]{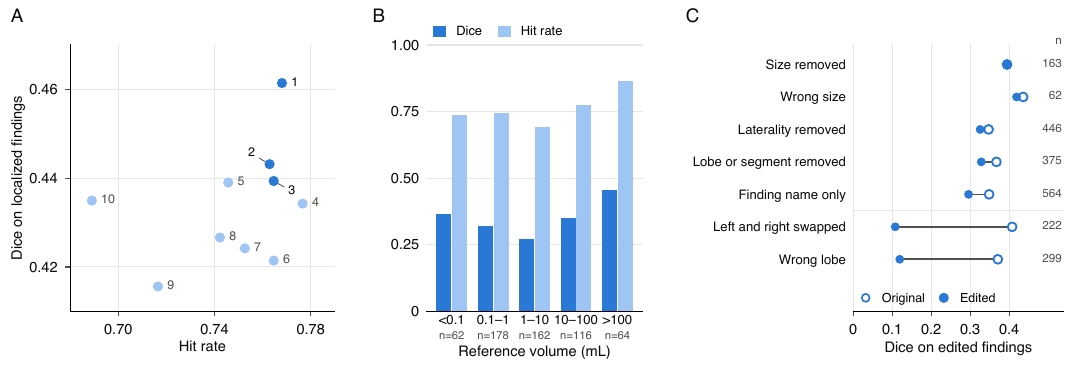}
\caption{Detection, finding size and prompts. A) Hit rate against mean Dice on localized findings
(Dice $\geq$ 0.1) for the ten best teams. Numbers give the final rank. The
three winning teams are shown in darker blue. B) Dice (dark) and hit rate
(light) by reference volume of the finding, pooled over the ten best teams.
C) Prompt sensitivity analysis. Dice on the findings changed by each edit, with
the original sentence (open) and the edited sentence (filled), averaged over
Teams 1 to 4. Numbers give the number of edited findings.}
\label{fig:further}
\end{figure*}

\subsection{Detection, finding size and prompts}
\label{sec:further}
Three analyses examine what separated the teams and which findings
were hard. First, detection and delineation. Across all 48 teams, hit
rate and Dice were closely related (Spearman $\rho = 0.97$), so teams
that localized more findings also segmented them better. Among the top four
teams, however, hit rates were almost identical (0.76--0.78), and the ranking
followed how well the localized findings were delineated. Team 1 reached a mean
Dice of 0.46 on the findings it localized, against 0.43--0.44 for Teams 2 to 4
(Figure~\ref{fig:further}A). Team 4 had the highest hit rate of all teams
(0.78) but the lowest Dice on localized findings of the four (0.43), which left
it fourth. Among focal findings, those with more than one reference instance had a
lower Dice than single-instance findings for all ten teams (0.34 against 0.37 on
average, 244 and 265 findings, respectively).

Second, finding size. Although small findings are usually the hardest to
segment, the ten best teams did not perform worst on the smallest findings. Nodules below 0.1\,mL reached a Dice of
0.39 and a hit rate of 0.78, similar to nodules of
0.1--1\,mL (Dice 0.38, hit rate 0.85). The lowest scores were for findings
of 1--10\,mL (Dice 0.27, hit rate 0.69), a range dominated
by linear opacities, and the highest for findings above 100\,mL
(Dice 0.46, hit rate 0.86), mostly ground-glass opacities, consolidations and
pleural effusions (Figure~\ref{fig:further}B). Size and finding type are
therefore entangled, and the dip reflects the difficulty of thin linear
opacities more than volume itself.

Third, prompts. Teams 1 to 4 ran their final methods on edited sentences
(Section~\ref{sec:prompt-exp}, Figure~\ref{fig:further}C). Removing the stated size changed Dice by less than 0.01 for all four teams (163
findings). For Teams 2, 3 and 4, removing the side lowered Dice by 0.02--0.03,
removing the lobe or segment by 0.04--0.06, and keeping only the name of the
finding by 0.05--0.08. Team 1 lost at most 0.02 under these edits and thus
relied less on the stated location. Wrong locations misled all four models.
With left and right swapped (examples in Figure~\ref{fig:prompt-examples}a and b), Dice fell from 0.40--0.42 to 0.08--0.14 and the
hit rate from 0.77--0.80 to 0.16--0.31, so the methods followed the location
given in the prompt. Teams 3 and 4 partly enforced this by post-processing, which
restricted masks to the named lobe or side, but Teams 1 and 2 used no such step,
so their models followed the location by themselves. Of the 222 findings whose side was swapped, Teams 2, 3 and 4
moved the mask to the named side (most of the mask on that side) for 60--65\%,
kept it on the side of the finding for 19--23\%, and returned an empty mask for
12--21\%. Team 1 moved the mask for 36\%, kept it for 33\% and returned an
empty mask for 31\%. Because the named side does not contain the finding, an
empty mask is arguably the most appropriate response, whereas moving the mask
means segmenting a finding that is not there. Naming the wrong lobe in the same lung lowered
Dice from 0.37--0.38 to 0.07--0.17 (examples in Figure~\ref{fig:prompt-examples}a--c).
Every edit that removed or changed the location lowered Dice significantly for
all four teams (Wilcoxon, adjusted $p < 0.001$). A
wrong size (example in Figure~\ref{fig:prompt-examples}a) left the masks of Teams 2, 3 and 4 almost unchanged (Dice changed
by less than 0.01), whereas the Dice of Team 1 fell from 0.46 to 0.41 on these
62 findings. Neither size edit changed Dice significantly for any team
(adjusted $p = 0.07$--0.89). When a finding in both lungs was prompted for one side only
($n = 297$ findings, example in Figure~\ref{fig:prompt-examples}c), at least 85\% of the predicted mask lay on the named side, and
the union of the two masks scored within 0.01 of the original sentence. The
models therefore relied on the stated side and lobe but hardly on the stated size.

\section{Discussion}
\label{sec:discussion}

\subsection{Interpretation of results}
The ReXGrounding Challenge was organized to advance the 3D grounding of
radiology report findings. Such grounding is a step towards tools that let
clinicians see, measure and track each reported finding in the image. It is also
a step towards the ultimate goal of grounded radiology report generation, in
which AI systems pair each finding they generate with its location in the volume
so that their output can be verified.

The challenge substantially advanced the state of the art. On the 100
scans of the pre-challenge benchmark, the winning team reached a Dice of
0.390, against 0.285 for the best model before the challenge. On the
full test set, the best entries reached a Dice of 0.375 and found up to
78\% of findings. Performance depended strongly on
the type of finding. Focal findings with a clear boundary, such as nodules,
consolidation and pleural effusion, were segmented far better than diffuse
non-focal findings. The low scores on non-focal findings partly reflect the
findings rather than the methods, because their extent is hard to define and
radiologists also disagree on it. The winner's margin nonetheless came mostly
from non-focal findings, in particular emphysema (Figure~\ref{fig:category-perf}).

Measured by Dice and hit rate, the best methods now match the agreement between
two independent groups of radiologists, who reached a Dice of 0.32 and a
hit rate of 0.76 when annotating the same 50 scans
(Section~\ref{sec:data}). The same holds for focal findings: focal Dice of the
top teams (0.35--0.38) is at or above inter-reader agreement on focal
findings (0.34), and non-focal Dice (0.18--0.24) is above the
0.15 reached by radiologists, although that figure rests on only 17
non-focal findings. The comparison is approximate, because the
inter-reader study used a different set of scans, but it suggests that Dice
against a single reference is close to saturating against annotation
variability.

The challenge moved this task forward substantially, and grounding report
findings in 3D chest CT is becoming feasible, at least for focal findings, which
make up most of the lung and pleural findings in chest CT reports (79\% of
training findings in the CT-RATE reports used here). A model that localizes a reported nodule or
consolidation could link each sentence of a report to the image, so that
referring physicians can find what the report describes, and could support
measurement and comparison over time. Even at the current level of performance,
such models could serve in tools that propose a region for a reader to confirm.
For method development, the challenge provides the first public benchmark of
this task at scale, with exhaustive radiologist annotation and a public
leaderboard that keeps the test masks hidden, so that new methods can be
compared directly with the entries reported here.

\subsection{Limitations and future work}
The task is not yet solved. Instance F1 remained well below inter-reader
agreement (0.34 for the best team against 0.47), so models often
found part of a multi-instance finding but not every lesion, and findings with
more than one instance had a lower Dice than single-instance findings
(Section~\ref{sec:further}). The qualitative failures also show that models
sometimes segmented the right abnormality at a location other than the one
stated in the sentence, which is the failure that grounding is meant to prevent. Finally, these results hold only for
non-contrast chest CT and for findings in the lungs and pleura, so grounding
in contrast-enhanced scans, in other body regions and for other types of
findings remains under-explored.

The challenge also has several limitations. All scans come from a single source,
CT-RATE, acquired at one centre, so the results may not transfer to other
institutions. The test set
is dominated by focal findings, non-focal findings make up only 12.5\% of
it, and several categories contain fewer than ten findings, so per-category
results for rare findings are imprecise. For ranking, Dice is a poor fit for diffuse
findings, whose extent varies even between radiologists. More broadly, agreement between radiologists was itself
modest (Dice 0.32 on the same 50 scans), because the annotation
protocol left room for interpretation, for example in how far a diffuse finding
extends or which of several lesions a sentence refers to. A single reference
mask is therefore an uncertain target, and small differences between teams are
hard to interpret. The training masks cover at most three instances of each
finding whereas the test masks are exhaustive, which penalizes models for
instances that the training data left unlabelled. Because the test scans and
their reports are publicly available in CT-RATE, models pretrained on CT-RATE
may have seen the test images and reports, although never the reference masks.

Future work could address these limitations in four ways. First, evaluation
sets for this task should use standardized
definitions for non-focal findings, with explicit rules for where a diffuse
finding ends and which instances a sentence covers, and should have each test
finding segmented by several independent readers. This would raise agreement
between readers, allow methods to be compared against a consensus reference,
and make scores on non-focal findings interpretable.
Second, data
from several centres, including contrast-enhanced CT, would test how well
methods generalize. Third, the task could be extended beyond chest CT and
beyond lung and pleural findings to other organs, body regions and imaging
modalities. Fourth, it could move from grounding given
findings to grounded report generation, in which a model both writes the
report and segments every finding it describes.

\section{Conclusion}
\label{sec:conclusion}
The ReXGrounding Challenge was the first challenge on segmenting findings
described in free-text radiology reports in 3D chest CT. The best
methods, built on pretrained text-promptable models with large anatomical
context, ensembling and anatomy-aware post-processing, reached the level of
agreement between independent radiologists on overall Dice. Three gaps remain.
Models often find only some of the lesions of a multi-instance finding, segment
diffuse findings far less accurately than focal ones, and sometimes place the
mask at a location other than the one the sentence names. These remaining gaps should guide the next generation of grounding methods. The public
leaderboard at \url{https://rexrank.ai/ReXGroundingCT} remains open, so that
new methods can be evaluated on the same hidden test set.

\section*{Declarations}
\paragraph{Author contributions}
Organizers. Mohammed Baharoon: Conceptualization, Methodology, Software, Formal analysis, Visualization, Project administration, Writing (original draft). Mahmoud Hussain Alabbad, Sung Eun Kim and Xiaoman Zhang: Project administration (challenge organization). Lucas Bijnens: Project administration (annotation organization). Luyang Luo: Project administration (challenge organization), Writing (review and editing). Pranav Rajpurkar: Supervision, Writing (review and editing).\\
Annotation. Michael Moritz and Hassan R. Alomaish: Data curation, Supervision. Ali Piracha, Reem Alwasiah, Talal Yahyan, Abdulrahman Alamoudi, Mashael Alshehri, Rayan Alsager, Hannah Hancock, Anna M. Hunkemöller, Peter Youssef, Mona A. Alhammad, Afnan Aljumayan, Brecht Van Berkel, Eric Bijnens, Saud Alhussain, Maarten Steyvers, Rolf Symons, Bryan Eelen, Ayed Alaklubi, Maha Buraykan Alshammari, Amirah Fahad Alshammeri, Tomas Gilvydis, Ibrahim AlTurki, Monerah Almohideb, Ali Alburkani, Shahad Alhomaiani and Fatima Jabali: Data curation.\\
Team~1. Zirong Chen: Data curation, Methodology, Software, Validation. Jiayu Wang: Formal analysis, Investigation, Methodology, Software. Xi Ouyang: Methodology, Software, Validation, Visualization. Xiaodan Ye: Data curation, Formal analysis, Investigation. Weiqiu Jin: Data curation, Formal analysis, Investigation. Mingyang Yu: Formal analysis, Investigation, Software. Dijia Wu: Formal analysis, Investigation, Methodology, Writing (review and editing).\\
Team~2. Kwang-Hyun Uhm: Conceptualization, Formal analysis, Investigation, Methodology, Project administration, Software, Supervision, Validation. Inhwa Son: Data curation, Formal analysis, Validation.\\
Team~3. Rachel van~'t Hull: Conceptualization, Data curation, Formal analysis, Investigation, Methodology, Software, Validation, Visualization, Writing (original draft). Tim Veenboer: Methodology, Resources. Cees Snoek: Supervision. Kevin Groot Lipman: Conceptualization, Methodology, Resources, Software, Supervision, Writing (review and editing).\\
Team~4. Daesung Kim: Conceptualization, Data curation, Formal analysis, Investigation, Methodology, Project administration. Junha Im: Data curation, Formal analysis, Investigation. Youngjun Jun: Formal analysis, Investigation. Sangwon Lee: Funding acquisition (Team 4). Katelyn Morrison: Conceptualization, Supervision. Adam Perer: Funding acquisition (Team 4), Supervision. Mijin Yun: Funding acquisition (Team 4), Supervision.\\
Maximilian Rokuss, Moritz Langenberg, Yannick Kirchhoff, Benjamin Hamm, Klaus Maier-Hein, Ibrahim Ethem Hamamci and Sezgin Er: Writing (review and editing).
\paragraph{Conflicts of interest} Zirong Chen, Jiayu Wang, Xi Ouyang and Dijia
Wu are employees of Shanghai United Imaging Intelligence, Shanghai, China,
where they work on research and development of AI algorithms for medical
imaging, and Mingyang Yu was an intern there during this work. Kwang-Hyun Uhm is Chief Technology Officer
of and holds equity in MEDAI Inc., a company developing medical imaging AI
software. MEDAI Inc. provided no funding for this work. Mijin Yun is the Chief
Technology Officer and Daesung Kim is an employee of NewcureM Inc., a medical AI
and radiopharmaceutical company. The PhD positions of Rachel van~'t
Hull and Tim Veenboer at the Netherlands Cancer Institute are funded by Kaiko,
a company developing medical imaging foundation models, through a collaboration
with the institute, and Kevin Groot Lipman is employed in part by Kaiko. Kaiko
was not otherwise involved in this work. Ibrahim Ethem Hamamci and Sezgin Er are
employees of Forithmus, a company developing AI for radiology and medical imaging. The remaining authors declare no
conflicts of interest.
\paragraph{Data availability} The ReXGroundingCT findings and reference masks
for the training and validation sets, and the findings for the test set, are
available at
\url{https://huggingface.co/datasets/rajpurkarlab/ReXGroundingCT} under a
CC BY-NC-SA 4.0 licence. The CT scans and reports are available from CT-RATE
\citep{hamamci2026generalist}. The test-set reference masks remain withheld so
that the public leaderboard at \url{https://rexrank.ai/ReXGroundingCT} can
continue to evaluate new methods on the same test set.
\paragraph{Code availability} The evaluation code that produced the rankings
is available with the dataset at
\url{https://huggingface.co/datasets/rajpurkarlab/ReXGroundingCT}. The code of
Team~1 is
available at \url{https://github.com/uii-open-source/CIRCLE/}, and its model
weights are not publicly released. The code of Team~2 will be released at
\url{https://github.com/khuhm/Grounder}. The code of Team~3 is available at
\url{https://github.com/nki-radiology/TAGS-ReXGroundingCT} under the Apache-2.0
licence, and its model weights at
\url{https://huggingface.co/fomofo/TAGS-ReXGroundingCT} (CC BY-NC-SA 4.0 for the
VoxTell models and CC BY-NC 4.0 for the TAP-CT models). The code of Team~4 is
available from the team on request.
\paragraph{Funding} The challenge and the preparation of this article did not
receive any specific grant from funding agencies in the public, commercial, or
not-for-profit sectors. Funding acquired by individual participating teams
supported only their own entries.

\section*{Declaration of generative AI and AI-assisted technologies in the manuscript preparation process}
During the preparation of this work the authors used Claude (Opus 5.5 and
Fable 5.1, Anthropic) to help draft and edit the text, write analysis and figure code, and check reported
numbers against the analysis outputs. After using this tool, the authors
reviewed and edited the content as needed and take full responsibility for the
content of the published article.

\bibliographystyle{elsarticle-harv}
\bibliography{references}

\appendix
\renewcommand{\theHtable}{app.\thetable}\renewcommand{\theHfigure}{app.\thefigure}

\section{Challenge organization}
\label{app:organization}

\paragraph{Name and organizers}
The ReXGrounding Challenge was held as part of MICCAI 2026. It was organized
by Mohammed Baharoon, Pranav Rajpurkar, Luyang Luo, Xiaoman Zhang, and
Sung Eun Kim (Harvard Medical School, USA) and Mahmoud Hussain Alabbad (King
Fahad Hospital, Saudi Arabia). The challenge design was registered in advance
with MICCAI and is publicly available \citep{baharoon2026design}.

\paragraph{Life cycle and platform}
The challenge was a one-time event with a fixed submission deadline of
14 September 2026, 11:59 PM ET, at which the official rankings were
fixed. It was hosted on the ReXrank leaderboard platform
\citep{zhang2024rexrank}
(\url{https://rexrank.ai/ReXGroundingCT}), where teams registered, submitted
predictions for the test set, and followed their scores on the leaderboard half of the
test set during the development phase. Results were announced at a dedicated
session at MICCAI 2026 in Strasbourg, France, on 27 September 2026, after which
the leaderboard reopened for new submissions as an ongoing benchmark.

\paragraph{Participation, awards and publication}
Members of the organizers' institutions could participate and were eligible
for awards, although none took part. The six best-performing teams received
certificates and presented their methods at the challenge session at MICCAI
2026 in Strasbourg, France, on 27 September 2026. Co-authorship of this
paper was originally offered to the top three teams. Because the fourth-placed
team finished only 0.0016 Dice behind the third, a margin too small to
separate them meaningfully, we extended co-authorship to the top four
teams (up to eight members per team). Teams were free to
publish their own results with no embargo.

\paragraph{Submission}
Teams registered on the ReXrank leaderboard and submitted predictions as a link
to a single ZIP archive on Google Drive. The archive held one NIfTI file per
test scan, with one binary mask channel per finding in the order given in the
test file. Submissions were downloaded and scored automatically every two
hours, and the leaderboard showed each submission's scores on the leaderboard half
of the test set. There was no separate test phase and no limit on the number
of submissions, but teams were asked not to submit again until their previous
submission had been scored. Teams made between 1 and 242
submissions (median 4). The submission used for the final ranking is
described in Section~\ref{sec:assessment}.

\paragraph{Timeline}
The ReXGroundingCT training set and original validation scans had been public
since 28 July 2025, and the 150 additional validation scans were
released on 13 June 2026. Registration and submission opened on
14 June 2026, and the first submission was scored on 2 July 2026.
By the deadline of 14 September 2026, 11:59 PM ET,
120 teams had registered. Results were announced at the MICCAI 2026 session on 27 September 2026.

\paragraph{Ethics, data use, code, and conflicts of interest}
All scans come from CT-RATE, a public dataset of de-identified chest CT scans
released under its own ethics approval \citep{hamamci2026generalist}, so
no additional approval was required for the challenge. The ReXGroundingCT data,
including the annotations used in the challenge, are distributed under a
CC BY-NC-SA 4.0 licence. The evaluation code that produced the rankings is
public with the dataset (\url{https://huggingface.co/datasets/rajpurkarlab/ReXGroundingCT}).
Teams were encouraged but not required to release their code, and the top teams
were required to describe their methods in sufficient detail for this paper. The
challenge had no sponsors or prizes. Access to the test-set reference masks was
restricted to members of the Rajpurkar Lab at Harvard Medical School, who were not
eligible to participate.

\section{Participating team method descriptions}
\label{app:methods}
\setcounter{figure}{0}
Each of the four winning teams described its method using a common template
whose headings follow the BIAS guideline. We edited the descriptions lightly
for length and style, and report scores on the full test set.

\subsection{Team 1}
\label{app:team1}

\paragraph{Data} Segmentation training used the 2,992 ReXGroundingCT
training scans with their findings and masks. The image encoder, CIRCLE, was
pretrained on 403,500 non-contrast chest CT scans from 288,740
patients, paired with radiology reports. Additional in-house data comprised
950 scans from the CIRCLE cohort at Zhongshan Hospital, Fudan University,
Shanghai, China: 800 scans with pulmonary nodules and 150 challenging
scans covering lung cancer, pneumonia and other infections, and less common
thoracic conditions. Positive finding descriptions were extracted from each
report, and annotation engineers with medical backgrounds exhaustively
delineated every visible abnormality corresponding to each finding, yielding
paired descriptions and 3D masks. No test-set images or text were used at any
stage, including the pretraining of components.

\paragraph{Pre-processing} Volumes were resampled to
0.65 $\times$ 0.65 $\times$ 1.15\,mm, with an input size of
384 $\times$ 320 $\times$ 224 voxels.

\begin{figure}[!htbp]
\centering
\includegraphics[width=\columnwidth]{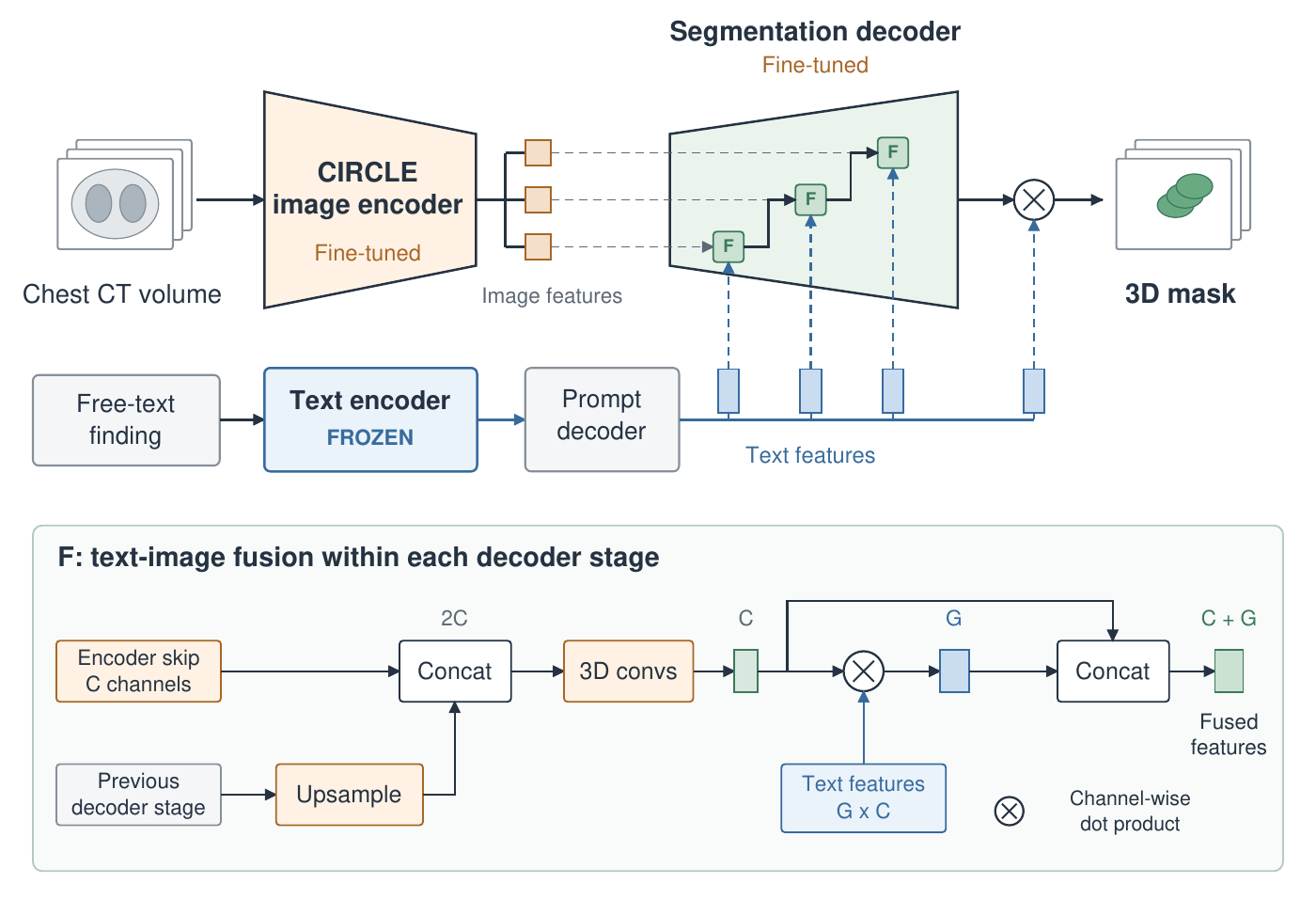}
\caption{Team 1: the pretrained CIRCLE image encoder combined with a VoxTell-based
decoder. The image encoder and segmentation decoder are fine-tuned, and the text
encoder is frozen. Inset F shows the text--image fusion within each decoder
stage.}
\label{fig:app-team1}
\end{figure}

\paragraph{Architecture and text encoder} CIRCLE pretraining jointly learns
image--report alignment and abnormality recognition. A 3D adaptation of
EfficientNetV2-XL \citep{tan2021efficientnetv2} and a Chinese BERT-based report
encoder project CT volumes and reports into a shared 512-dimensional space. A
SigLIP-style contrastive objective \citep{zhai2023sigmoid} aligns
image--report pairs, while a parallel classification head predicts 37 thoracic
abnormalities from report-derived labels with binary cross-entropy. For
grounding, the pretrained image encoder was integrated into VoxTell
\citep{rokuss2026voxtell} (Figure~\ref{fig:app-team1}): a prompt decoder
conditions segmentation on features from a frozen Qwen3-Embedding-4B text encoder
\citep{zhang2025qwen3}, and the image encoder and segmentation decoder are
fine-tuned.

\paragraph{Training} Dice plus binary cross-entropy loss, AdamW with an initial
learning rate of $3 \times 10^{-4}$, and a batch size of 24. Training took about
two days on eight H20 GPUs.

\paragraph{Ensembling} Probability maps from checkpoints of different epochs
were averaged voxel-wise.

\paragraph{Post-processing} The averaged probability map was thresholded at
0.5. No other post-processing was applied.

\paragraph{Inference time and hardware} About 20\,s per scan per finding on one
H20 GPU.

\paragraph{Final submission and lessons} The final submission is listed as
``CIRCLE'' on the leaderboard. On a 50-case validation subset, pretraining
raised mean Dice from 0.277 with random initialization to 0.327,
against 0.282 for VoxTell, and the in-house data raised it further to
0.356. Pretraining and additional annotations therefore helped, whereas the
same architecture trained from scratch did not improve on VoxTell.

\paragraph{Code and weights} Code: \url{https://github.com/uii-open-source/CIRCLE/}.
Model weights are not publicly released.

\subsection{Team 2}
\label{app:team2}

\paragraph{Data} The challenge training set (2,992 scans with findings and
masks), with the 200-scan validation set used only for model selection.
Annotations were used unmodified. No external or in-house CT data, pseudo-labels
or derived anatomy masks were used in the final system. The only externally
pretrained component was Qwen3-Embedding-4B \citep{zhang2025qwen3} (Apache 2.0),
used frozen, and the image network was trained from scratch. Test-set scans and
finding texts were not used for training or model selection. Whether any such
material was part of the text encoder's own pretraining data is not known.

\paragraph{Pre-processing} Following nnU-Net preprocessing \citep{isensee2021nnu},
volumes were oriented to LPS and resampled to about
0.7 $\times$ 0.7 $\times$ 1.0\,mm. Intensities were clipped to
$[-1000, 197]$\,HU and normalized with a mean of $-549$\,HU and a standard
deviation of 313\,HU. Training used random foreground-centred crops with zero
padding, with patch sizes from 224 $\times$ 224 $\times$ 128 to
448 $\times$ 448 $\times$ 256 voxels.

\paragraph{Architecture and text encoder} The model, Grounder, uses a 3D nnU-Net
residual-encoder (ResEnc-L) encoder \citep{isensee2024nnu} and a five-stage
decoder. The frozen Qwen3-Embedding-4B converts each finding into a
2,560-dimensional embedding, and two-layer MLPs generate scale and shift
parameters that condition the decoder features through FiLM
\citep{perez2018film}, $F' = \gamma F + \beta$.

\paragraph{Training} Dice plus cross-entropy loss, AdamW with a fixed learning
rate of $10^{-4}$, and a batch size of one. Models generally converged within
20--30 epochs. Image augmentation included rotation, scaling and intensity
transformations, but no flipping. In-distribution text augmentation replaced a
finding with a real finding from another training case that matched it in
location and finding type, keeping the original CT and mask. Each model trained
on one H200 GPU for about 4--7 days, depending on patch size.

\paragraph{Ensembling} Thirteen independently trained models differing in patch
size, selected on validation performance, with voxel-wise probabilities averaged
equally.

\paragraph{Post-processing} Probabilities were thresholded at 0.5. No
connected-component filtering, anatomical restriction, or rewriting of the text
at inference was applied.

\paragraph{Inference time and hardware} About 5\,s per scan per finding on one
H200 GPU.

\paragraph{Final submission and lessons} The final submission was
Grounder\_1.4n (full test set: Dice 0.365, hit rate 0.763, instance
F1 0.343). Larger anatomical context, in-distribution text augmentation
and diverse patch sizes improved grounding. Adding lobe maps from
TotalSegmentator \citep{wasserthal2023totalsegmentator} or category embeddings
did not help, and augmentation with negative text was unreliable because the
training annotations are partial.

\paragraph{Code and weights} The repository \url{https://github.com/khuhm/Grounder}
(MIT licence) will hold the code and trained weights, which have not yet been
released.

\subsection{Team 3}
\label{app:team3}

\paragraph{Data} The challenge training (2,992 scans) and validation
(200 scans) sets. External data were 8,000 sentence--mask pairs from
CT-RATE \citep{hamamci2026generalist}, with pseudo-masks of nodules and effusions
from TotalSegmentator \citep{wasserthal2023totalsegmentator}, used by two
ensemble members (one of them only the 1,540 effusion pairs). In-house data
were used only for the pretraining of TAP-CT \citep{veenboer2025tap}
(105,000 CT volumes). Pretrained weights were VoxTell v1.1
\citep{rokuss2026voxtell}, Qwen3-Embedding \citep{zhang2025qwen3}, TAP-CT-B-3D
and Qwen2.5-7B \citep{yang2025qwen3}. Derived data were TotalSegmentator lobe
masks, a lobe mapping of all finding sentences by Qwen3-32B
\citep{yang2025qwen3}, synonym-based text augmentation and Qwen2.5 paraphrases.
Test-set exposure: the CT-RATE pool incidentally included test studies, so one
member saw 80 of the 300 test studies and the other member 10 of these,
as image, report sentence and pseudo-mask, but never the challenge masks or
sentences. Retraining both members with these studies replaced by other CT-RATE
scans changed the final masks equally on exposed and unexposed studies (Dice
0.976 vs.\ 0.974 between versions).

\paragraph{Pre-processing} VoxTell members use the native grid, z-scored
intensities and random $192^3$ crops, one third of them centred on a finding.
TAP-CT resamples to 1.5\,mm isotropic voxels, clips to $[-1008, 822]$\,HU,
z-scores, and uses $24 \times 224 \times 224$ crops.

\paragraph{Architecture and text encoder} VoxTell: an nnU-Net residual encoder
\citep{isensee2021nnu,isensee2024nnu} and a transformer decoder in which the
frozen Qwen3-Embedding-4B sentence embedding queries the $12^3$ bottleneck tokens
(440M trainable parameters). TAP-CT: a 3D ViT-B (87M parameters) conditioned on
Qwen3-Embedding-0.6B through FiLM \citep{perez2018film}, followed by a
convolutional decoder.

\paragraph{Training} VoxTell members were trained from VoxTell v1.1 for 250
epochs, five of them for up to 430, with deeply supervised Dice plus
cross-entropy loss (Tversky loss in three members), SGD at $10^{-4}$ with
polynomial decay, and a batch size of 8, taking about 7\,h per 250 epochs on one
H100. There was no image augmentation, and text augmentation replaced phrases
with equivalent phrasings seen in the data ($p = 0.5$). The validation data were
in every member's training set. TAP-CT members used Dice plus cross-entropy,
AdamW and a batch size of 16, with 50,000 steps with the encoder frozen followed
by 25,000 steps of fine-tuning (4.5\,h per seed on one H100), five seeds,
left--right flips, paraphrases and negatives with swapped locations.

\begin{figure*}[!htbp]
\centering
\includegraphics[width=\textwidth]{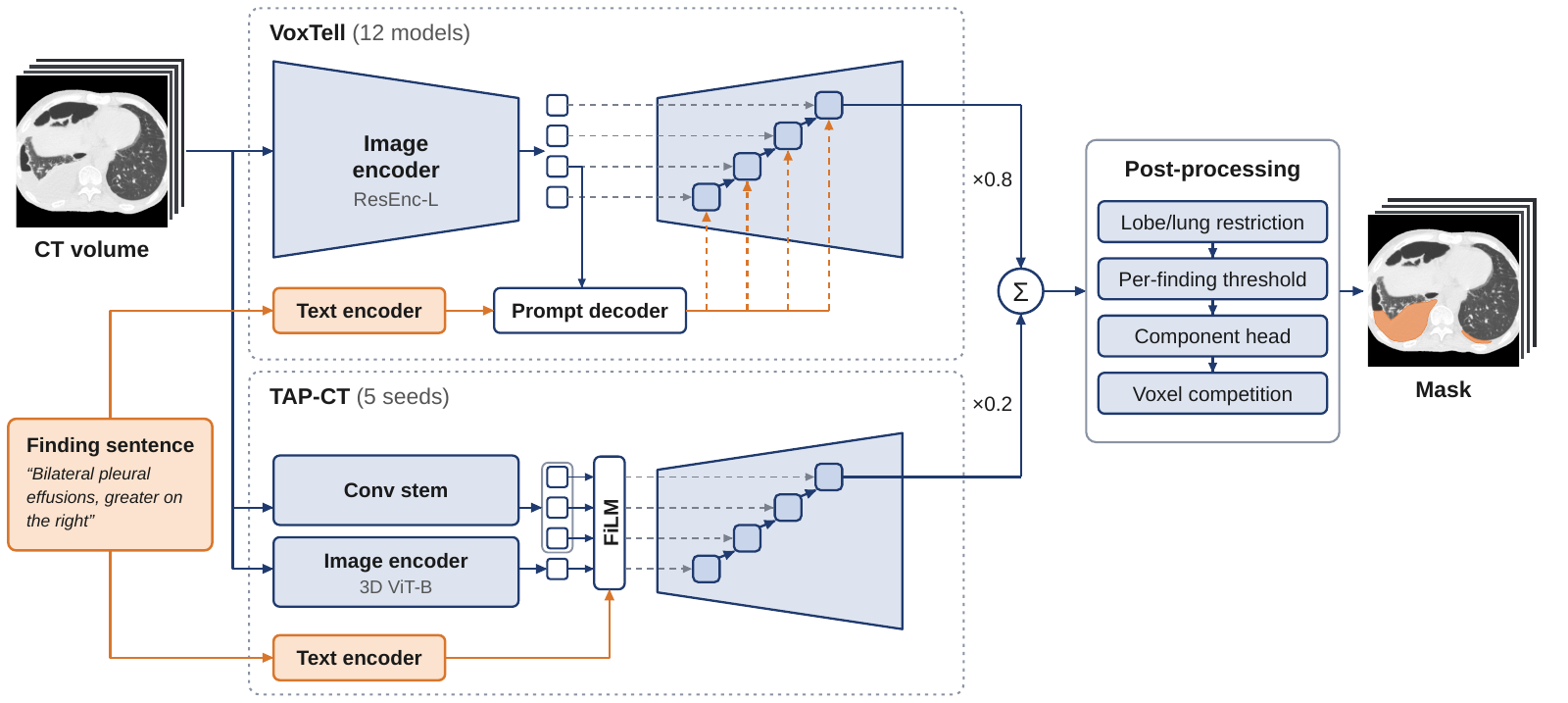}
\caption{Team 3: twelve VoxTell models and five TAP-CT seeds, each conditioned on
the finding sentence, produce probability maps that are averaged per branch and
fused 0.8/0.2, followed by four post-processing steps.}
\label{fig:app-team3}
\end{figure*}

\paragraph{Ensembling} The probability maps of twelve VoxTell members, which
differ in initialization, loss, sampling and auxiliary data, are averaged and
fused 0.8/0.2 with the mean of five TAP-CT seeds (Figure~\ref{fig:app-team3}).

\paragraph{Post-processing} The fused map is restricted to the named lobe
(dilated by 16 voxels) or lung (12 voxels), thresholded per finding by a
gradient-boosted regressor, and cleaned by a small 3D CNN that removes components
predicted to lower Dice (both fitted on the validation set). Voxels claimed by
several findings go to the most confident one. The only language model used at
inference is the Qwen3-32B lobe mapping.

\paragraph{Inference time and hardware} On one H100, a VoxTell member takes
3.7\,min per scan for all findings jointly, about 24 GPU-minutes per finding for
the twelve members, and the five TAP-CT seeds together take 23\,s per scan.

\paragraph{Final submission and lessons} The final submission was TAGS\_TAPv3
(full test set Dice 0.362). Most helpful were TAP-CT as a second
architecture, the dilated lobe restriction and the per-finding thresholds.
Changes targeting location words, pseudo-labels and extra public data did not
help. Four late post-processing additions each gained 0.001--0.004 on the
validation set and the leaderboard half, yet the submission without them scored
0.365 on the full test set.

\paragraph{Code and weights} Code:
\url{https://github.com/nki-radiology/TAGS-ReXGroundingCT} (Apache-2.0).
Weights of all members: \url{https://huggingface.co/fomofo/TAGS-ReXGroundingCT}
(CC BY-NC-SA 4.0, TAP-CT seeds CC BY-NC 4.0).

\subsection{Team 4}
\label{app:team4}

\paragraph{Data} The challenge training data and 200 validation scans,
COVID-19-20 (180 scans) \citep{roth2022rapid},
OmniAbnorm-CT (52) \citep{zhao2026rethinking}, a sparse interstitial lung disease
dataset (91) \citep{depeursinge2012building}, and 18,652 CT-RATE
scans \citep{hamamci2026generalist}, excluding the test-set scans, with PatchChestCT
annotations \citep{li2026patchchestct} for 768 of them. The team reviewed
the training annotations and edited 146 findings in 117 scans. Common
errors were drawing artefacts, partly caused by the annotation tool, mismatches
between the location in the sentence and the mask, incomplete annotations such
as missed nodules, arteries annotated as atelectasis or fibrosis, and the whole
lung drawn as emphysema.

\paragraph{Pre-processing} Volumes were resampled to
0.75 $\times$ 0.75 $\times$ 1.25\,mm and cropped to the lungs using
TotalSegmentator lung masks \citep{wasserthal2023totalsegmentator} with 15\,mm
margins. Intensities were clipped to $[-1024, 1000]$\,HU and normalized as
$(\mathrm{HU} + 232)/281$.

\paragraph{Architecture and text encoder} VoxTell v1.1 \citep{rokuss2026voxtell}
with frozen Qwen3-Embedding-4B embeddings \citep{zhang2025qwen3} and input crops
of 288 $\times$ 288 $\times$ 192 voxels.

\begin{figure}[!htbp]
\centering
\includegraphics[width=\columnwidth]{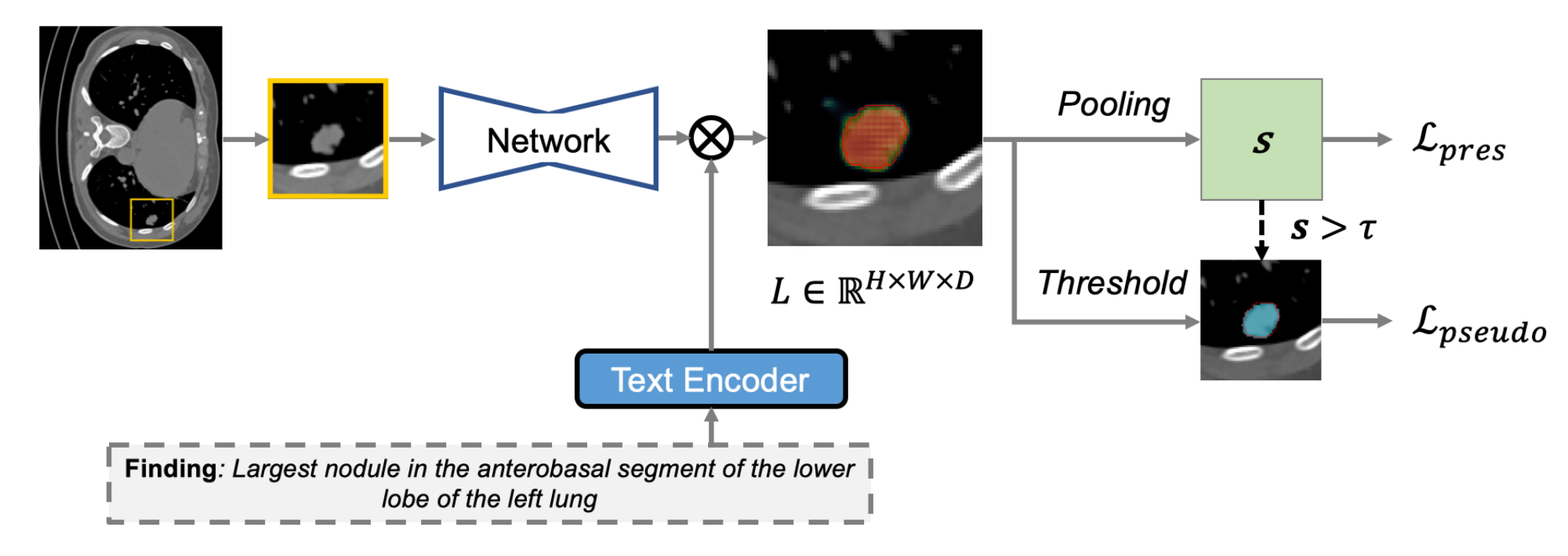}
\caption{Team 4: semi-supervised training on CT-RATE (stage 2). Voxel logits for
the finding are pooled into a presence score trained against whether the finding
is present, and thresholded predictions of an EMA teacher serve as pseudo-masks.}
\label{fig:app-team4}
\end{figure}

\paragraph{Training} Models were trained with an effective batch size of 8
across two RTX PRO 6000 GPUs, using AdamW with an initial learning rate of
$10^{-5}$ and cosine decay unless stated otherwise, in three stages. First, two
models were initialized from VoxTell v1.1 and trained on the challenge training
set and complementary halves of the validation set, with the other half kept for
checkpoint selection. The loss combined 0.3 binary cross-entropy, 0.7 Dice and
0.1 location KL divergence with deep supervision, and augmentation included
intensity changes, noise, sharpening and paraphrased prompts. Second, training
continued with the external segmentation datasets and with semi-supervised
training on CT-RATE (Figure~\ref{fig:app-team4}). Positive findings were
extracted from the CT-RATE reports with MedGemma \citep{sellergren2026open}, and
two losses were added: a presence loss (weight 0.1), in which voxel logits are
pooled with log-sum-exp and compared with whether the finding is present by
binary cross-entropy, and a pseudo-mask loss from an EMA teacher (weight 0.2,
decay 0.995). Third, each selected model was fine-tuned on the full validation
set alone, with the supervised losses, for 500 steps at a constant learning rate
of $3 \times 10^{-7}$.

\paragraph{Ensembling} The two models from the two validation folds were
combined by averaging their probabilities.

\paragraph{Post-processing} For findings that name the left or right side,
predictions confidently assigned to a lobe on the wrong side, according to
TotalSegmentator, were removed.

\paragraph{Inference time and hardware} 5.90\,s per finding
(11.45\,s per scan) on an RTX 3090, excluding input, output and
post-processing.

\paragraph{Final submission and lessons} The final submission was the ensemble
described above. Larger crops and fine-tuning on the curated data helped.
Re-annotating non-focal categories such as bronchiectasis, and a Tversky loss to
address the partial annotation of the training set, did not.

\paragraph{Code and weights} The edited data, code and weights will not be
publicly released, but are available on request.

\section{Full leaderboard}
\label{app:leaderboard}
\setcounter{table}{0}\setcounter{figure}{0}
Table~\ref{tab:leaderboard-full} lists all 48 ranked teams, and
Figure~\ref{fig:ranking} shows how stable the ranking was under bootstrap
resampling of the test scans.
\begin{table*}[!p]
\centering
\caption{Full leaderboard on the 300-scan test set, ordered by mean Dice.
Teams are numbered by their final rank.
Brackets give 95\% bootstrap confidence intervals (1,000
replicates, Section~\ref{sec:stability}). Submissions counts all submissions a
team made to the leaderboard. Institutions are those of all team members and countries as given at
registration, and -- marks information that was not provided.}
\label{tab:leaderboard-full}
\renewcommand{\baselinestretch}{1}\scriptsize
\setlength{\tabcolsep}{3pt}
\renewcommand{\arraystretch}{1.0}
\begin{tabularx}{\textwidth}{l>{\raggedright\arraybackslash}Xlcccccr}
\toprule
&&&&& \multicolumn{3}{c}{Instance} & \\
\cmidrule(lr){6-8}
Team & Institution & Country & Dice & Hit rate & Precision & Recall & F1 & Submissions \\
\midrule
Team 1 & United Imaging Intelligence, ShanghaiTech University, Zhongshan Hospital & China & 0.3755 {\tiny[0.352--0.399]} & 0.7680 {\tiny[0.732--0.802]} & 0.2720 {\tiny[0.242--0.300]} & 0.4570 {\tiny[0.413--0.502]} & 0.3410 {\tiny[0.312--0.369]} & 6 \\
Team 2 & Gachon University, MEDAI & South Korea & 0.3654 {\tiny[0.344--0.387]} & 0.7629 {\tiny[0.730--0.797]} & 0.2797 {\tiny[0.251--0.312]} & 0.4427 {\tiny[0.405--0.482]} & 0.3428 {\tiny[0.316--0.371]} & 19 \\
Team 3 & Netherlands Cancer Institute & Netherlands & 0.3621 {\tiny[0.342--0.383]} & 0.7646 {\tiny[0.737--0.795]} & 0.2537 {\tiny[0.226--0.287]} & 0.4869 {\tiny[0.451--0.522]} & 0.3336 {\tiny[0.306--0.364]} & 38 \\
Team 4 & Yonsei University, NewcureM & South Korea & 0.3605 {\tiny[0.339--0.382]} & 0.7766 {\tiny[0.744--0.809]} & 0.2419 {\tiny[0.215--0.273]} & 0.4377 {\tiny[0.398--0.477]} & 0.3116 {\tiny[0.285--0.341]} & 62 \\
\midrule
Team 5 & Deepnoid & South Korea & 0.3526 {\tiny[0.330--0.375]} & 0.7457 {\tiny[0.713--0.781]} & 0.1174 {\tiny[0.100--0.137]} & 0.4969 {\tiny[0.460--0.533]} & 0.1899 {\tiny[0.166--0.216]} & 21 \\
Team 6 & Siemens Healthineers, UCSF & USA & 0.3519 {\tiny[0.332--0.373]} & 0.7646 {\tiny[0.732--0.799]} & 0.1618 {\tiny[0.140--0.186]} & 0.4919 {\tiny[0.456--0.529]} & 0.2435 {\tiny[0.217--0.271]} & 242 \\
Team 7 & Raidium & France & 0.3432 {\tiny[0.322--0.366]} & 0.7526 {\tiny[0.721--0.785]} & 0.1829 {\tiny[0.162--0.208]} & 0.4558 {\tiny[0.422--0.492]} & 0.2611 {\tiny[0.238--0.288]} & 4 \\
Team 8 & German Cancer Research Center (DKFZ) & Germany & 0.3393 {\tiny[0.317--0.362]} & 0.7423 {\tiny[0.708--0.777]} & 0.1687 {\tiny[0.146--0.194]} & 0.4558 {\tiny[0.422--0.492]} & 0.2463 {\tiny[0.220--0.275]} & 7 \\
Team 9 & Shanghai Jiao Tong University & China & 0.3329 {\tiny[0.313--0.354]} & 0.7165 {\tiny[0.683--0.750]} & 0.1691 {\tiny[0.148--0.195]} & 0.4658 {\tiny[0.431--0.501]} & 0.2481 {\tiny[0.222--0.278]} & 12 \\
Team 10 & Hamad Bin Khalifa University & Qatar & 0.3329 {\tiny[0.311--0.355]} & 0.6890 {\tiny[0.654--0.726]} & 0.2313 {\tiny[0.202--0.263]} & 0.4209 {\tiny[0.384--0.456]} & 0.2985 {\tiny[0.269--0.328]} & 44 \\
\midrule
Team 11 & Friedrich-Alexander-Universit\"at Erlangen-N\"urnberg & Germany & 0.3317 {\tiny[0.311--0.353]} & 0.7406 {\tiny[0.706--0.774]} & 0.1225 {\tiny[0.105--0.142]} & 0.5044 {\tiny[0.469--0.537]} & 0.1971 {\tiny[0.172--0.223]} & 112 \\
Team 12 & Microsoft Research India & India & 0.3292 {\tiny[0.308--0.352]} & 0.7388 {\tiny[0.704--0.774]} & 0.1678 {\tiny[0.145--0.194]} & 0.4340 {\tiny[0.399--0.471]} & 0.2421 {\tiny[0.217--0.270]} & 5 \\
Team 13 & Shanghai Jiao Tong University, Shanghai Artificial Intelligence Laboratory, Medical Image Insights & China & 0.3272 {\tiny[0.307--0.348]} & 0.7148 {\tiny[0.680--0.751]} & 0.2662 {\tiny[0.237--0.298]} & 0.4191 {\tiny[0.382--0.460]} & 0.3256 {\tiny[0.298--0.355]} & 16 \\
Team 14 & Duke University & USA & 0.3262 {\tiny[0.305--0.349]} & 0.7371 {\tiny[0.703--0.770]} & 0.1460 {\tiny[0.127--0.169]} & 0.4533 {\tiny[0.420--0.488]} & 0.2208 {\tiny[0.198--0.248]} & 4 \\
Team 15 & Seoul National University & South Korea & 0.3258 {\tiny[0.306--0.346]} & 0.7251 {\tiny[0.689--0.760]} & 0.2179 {\tiny[0.193--0.245]} & 0.4054 {\tiny[0.371--0.444]} & 0.2835 {\tiny[0.259--0.309]} & 19 \\
Team 16 & Medical University of Vienna & Austria & 0.3234 {\tiny[0.305--0.343]} & 0.7491 {\tiny[0.714--0.779]} & 0.1629 {\tiny[0.141--0.187]} & 0.4558 {\tiny[0.424--0.488]} & 0.2400 {\tiny[0.214--0.268]} & 4 \\
Team 17 & Fudan University & China & 0.3231 {\tiny[0.302--0.343]} & 0.7354 {\tiny[0.700--0.771]} & 0.2858 {\tiny[0.257--0.315]} & 0.4203 {\tiny[0.381--0.459]} & 0.3402 {\tiny[0.313--0.366]} & 42 \\
Team 18 & University of Campinas & Brazil & 0.3203 {\tiny[0.299--0.340]} & 0.7131 {\tiny[0.675--0.749]} & 0.1217 {\tiny[0.104--0.142]} & 0.4857 {\tiny[0.444--0.522]} & 0.1946 {\tiny[0.170--0.221]} & 11 \\
Team 19 & LFAI & -- & 0.3202 {\tiny[0.299--0.340]} & 0.7165 {\tiny[0.682--0.752]} & 0.1063 {\tiny[0.091--0.125]} & 0.5093 {\tiny[0.469--0.546]} & 0.1759 {\tiny[0.153--0.202]} & 4 \\
Team 20 & Chongqing University, Hong Kong Polytechnic University, University of Massachusetts Lowell & China, USA & 0.3126 {\tiny[0.293--0.334]} & 0.7079 {\tiny[0.674--0.745]} & 0.1461 {\tiny[0.125--0.169]} & 0.4085 {\tiny[0.374--0.445]} & 0.2152 {\tiny[0.191--0.240]} & 3 \\
Team 21 & University of Ulsan, National Institutes of Health, Asan Medical Center & South Korea & 0.3085 {\tiny[0.287--0.330]} & 0.6821 {\tiny[0.643--0.725]} & 0.1429 {\tiny[0.124--0.163]} & 0.4452 {\tiny[0.410--0.479]} & 0.2164 {\tiny[0.192--0.241]} & 4 \\
Team 22 & Kosin University Gospel Hospital & South Korea & 0.2990 {\tiny[0.279--0.320]} & 0.7010 {\tiny[0.661--0.736]} & 0.1284 {\tiny[0.112--0.149]} & 0.3892 {\tiny[0.357--0.424]} & 0.1930 {\tiny[0.172--0.218]} & 12 \\
Team 23 & Muroran Institute of Technology & Japan & 0.2868 {\tiny[0.266--0.310]} & 0.6735 {\tiny[0.637--0.716]} & 0.1118 {\tiny[0.096--0.131]} & 0.3979 {\tiny[0.363--0.433]} & 0.1746 {\tiny[0.153--0.200]} & 3 \\
Team 24 & AMC & South Korea & 0.2855 {\tiny[0.264--0.307]} & 0.6495 {\tiny[0.614--0.687]} & 0.1130 {\tiny[0.098--0.130]} & 0.4402 {\tiny[0.402--0.478]} & 0.1799 {\tiny[0.159--0.203]} & 1 \\
Team 25 & Microsoft Research & UK & 0.2748 {\tiny[0.256--0.295]} & 0.6770 {\tiny[0.643--0.712]} & 0.1565 {\tiny[0.140--0.175]} & 0.3948 {\tiny[0.360--0.429]} & 0.2241 {\tiny[0.205--0.245]} & 4 \\
Team 26 & MBZUAI & UAE & 0.2737 {\tiny[0.252--0.297]} & 0.6014 {\tiny[0.563--0.639]} & 0.1653 {\tiny[0.143--0.192]} & 0.3039 {\tiny[0.272--0.336]} & 0.2141 {\tiny[0.191--0.239]} & 6 \\
Team 27 & Korea University & South Korea & 0.2693 {\tiny[0.248--0.289]} & 0.6615 {\tiny[0.623--0.702]} & 0.1258 {\tiny[0.110--0.145]} & 0.3960 {\tiny[0.363--0.431]} & 0.1909 {\tiny[0.171--0.213]} & 1 \\
Team 28 & Northwestern Polytechnical University & China & 0.2674 {\tiny[0.247--0.287]} & 0.6460 {\tiny[0.608--0.686]} & 0.0846 {\tiny[0.073--0.099]} & 0.4421 {\tiny[0.406--0.477]} & 0.1420 {\tiny[0.125--0.163]} & 8 \\
Team 29 & -- & -- & 0.2673 {\tiny[0.247--0.288]} & 0.6443 {\tiny[0.609--0.683]} & 0.2778 {\tiny[0.249--0.305]} & 0.3543 {\tiny[0.321--0.390]} & 0.3114 {\tiny[0.287--0.335]} & 3 \\
Team 30 & Robert Gordon University & UK & 0.2559 {\tiny[0.237--0.276]} & 0.6478 {\tiny[0.613--0.685]} & 0.1246 {\tiny[0.110--0.142]} & 0.3219 {\tiny[0.293--0.356]} & 0.1796 {\tiny[0.163--0.200]} & 2 \\
Team 31 & University of Birmingham & UK & 0.2492 {\tiny[0.226--0.273]} & 0.5447 {\tiny[0.504--0.588]} & 0.0948 {\tiny[0.081--0.112]} & 0.3014 {\tiny[0.270--0.335]} & 0.1442 {\tiny[0.127--0.165]} & 9 \\
Team 32 & University of Cambridge & UK & 0.2460 {\tiny[0.222--0.269]} & 0.5292 {\tiny[0.489--0.570]} & 0.0914 {\tiny[0.078--0.107]} & 0.2983 {\tiny[0.267--0.329]} & 0.1399 {\tiny[0.123--0.159]} & 1 \\
Team 33 & Chinese University of Hong Kong & Hong Kong SAR & 0.2301 {\tiny[0.208--0.252]} & 0.5172 {\tiny[0.477--0.559]} & 0.1102 {\tiny[0.093--0.132]} & 0.2964 {\tiny[0.262--0.334]} & 0.1606 {\tiny[0.140--0.186]} & 3 \\
Team 34 & Hangzhou Dianzi University, Tsinghua University & China & 0.2259 {\tiny[0.205--0.248]} & 0.5619 {\tiny[0.523--0.603]} & 0.1061 {\tiny[0.092--0.123]} & 0.2889 {\tiny[0.258--0.318]} & 0.1552 {\tiny[0.138--0.175]} & 1 \\
Team 35 & University of Birmingham & UK & 0.2209 {\tiny[0.201--0.242]} & 0.5069 {\tiny[0.462--0.551]} & 0.0837 {\tiny[0.072--0.097]} & 0.3176 {\tiny[0.285--0.351]} & 0.1325 {\tiny[0.117--0.150]} & 1 \\
Team 36 & Ant Group & China & 0.2201 {\tiny[0.200--0.243]} & 0.5103 {\tiny[0.473--0.550]} & 0.1048 {\tiny[0.091--0.123]} & 0.2771 {\tiny[0.246--0.311]} & 0.1520 {\tiny[0.134--0.173]} & 1 \\
Team 37 & Independent & -- & 0.2183 {\tiny[0.196--0.240]} & 0.4708 {\tiny[0.433--0.510]} & 0.0655 {\tiny[0.056--0.076]} & 0.2858 {\tiny[0.256--0.320]} & 0.1066 {\tiny[0.094--0.122]} & 1 \\
Team 38 & Asan Medical Center & South Korea & 0.2170 {\tiny[0.196--0.240]} & 0.5069 {\tiny[0.470--0.548]} & 0.0694 {\tiny[0.059--0.083]} & 0.2565 {\tiny[0.227--0.288]} & 0.1092 {\tiny[0.095--0.128]} & 2 \\
Team 39 & Pontificia Universidad Católica de Chile & Chile & 0.2136 {\tiny[0.192--0.235]} & 0.5241 {\tiny[0.485--0.566]} & 0.1032 {\tiny[0.088--0.123]} & 0.2746 {\tiny[0.243--0.306]} & 0.1500 {\tiny[0.132--0.172]} & 8 \\
Team 40 & UCSF & USA & 0.2098 {\tiny[0.189--0.233]} & 0.4845 {\tiny[0.448--0.525]} & 0.0801 {\tiny[0.068--0.096]} & 0.2503 {\tiny[0.222--0.283]} & 0.1214 {\tiny[0.106--0.141]} & 2 \\
Team 41 & Bangladesh University of Engineering and Technology & Bangladesh & 0.2096 {\tiny[0.189--0.233]} & 0.4863 {\tiny[0.448--0.526]} & 0.0805 {\tiny[0.069--0.096]} & 0.2503 {\tiny[0.222--0.283]} & 0.1218 {\tiny[0.107--0.142]} & 2 \\
Team 42 & Independent & South Korea & 0.1943 {\tiny[0.177--0.214]} & 0.5412 {\tiny[0.503--0.581]} & 0.0574 {\tiny[0.050--0.066]} & 0.3381 {\tiny[0.304--0.373]} & 0.0982 {\tiny[0.086--0.112]} & 2 \\
Team 43 & Erasmus University & Netherlands & 0.1936 {\tiny[0.174--0.214]} & 0.4948 {\tiny[0.457--0.533]} & 0.1913 {\tiny[0.166--0.222]} & 0.2210 {\tiny[0.193--0.250]} & 0.2051 {\tiny[0.182--0.230]} & 1 \\
Team 44 & University of Wisconsin--Madison & USA & 0.1834 {\tiny[0.165--0.203]} & 0.4897 {\tiny[0.453--0.530]} & 0.0833 {\tiny[0.069--0.100]} & 0.2522 {\tiny[0.218--0.288]} & 0.1252 {\tiny[0.105--0.147]} & 1 \\
Team 45 & Sichuan University of Science \& Engineering & China & 0.1787 {\tiny[0.159--0.198]} & 0.4003 {\tiny[0.364--0.437]} & 0.0549 {\tiny[0.047--0.064]} & 0.2335 {\tiny[0.206--0.263]} & 0.0890 {\tiny[0.078--0.101]} & 1 \\
Team 46 & Nagoya University & Japan & 0.1573 {\tiny[0.138--0.176]} & 0.3763 {\tiny[0.342--0.415]} & 0.0556 {\tiny[0.047--0.065]} & 0.2080 {\tiny[0.183--0.234]} & 0.0877 {\tiny[0.076--0.100]} & 1 \\
Team 47 & Bangladesh University of Engineering and Technology & Bangladesh & 0.1545 {\tiny[0.139--0.171]} & 0.4708 {\tiny[0.434--0.510]} & 0.1660 {\tiny[0.142--0.196]} & 0.1800 {\tiny[0.157--0.206]} & 0.1727 {\tiny[0.152--0.198]} & 2 \\
Team 48 & Technical University of Munich & Germany & 0.1461 {\tiny[0.130--0.162]} & 0.4192 {\tiny[0.378--0.463]} & 0.0294 {\tiny[0.024--0.035]} & 0.3481 {\tiny[0.312--0.384]} & 0.0542 {\tiny[0.045--0.064]} & 2 \\
\bottomrule
\end{tabularx}
\end{table*}

\begin{figure*}[!tp]
\centering
\includegraphics[width=\textwidth]{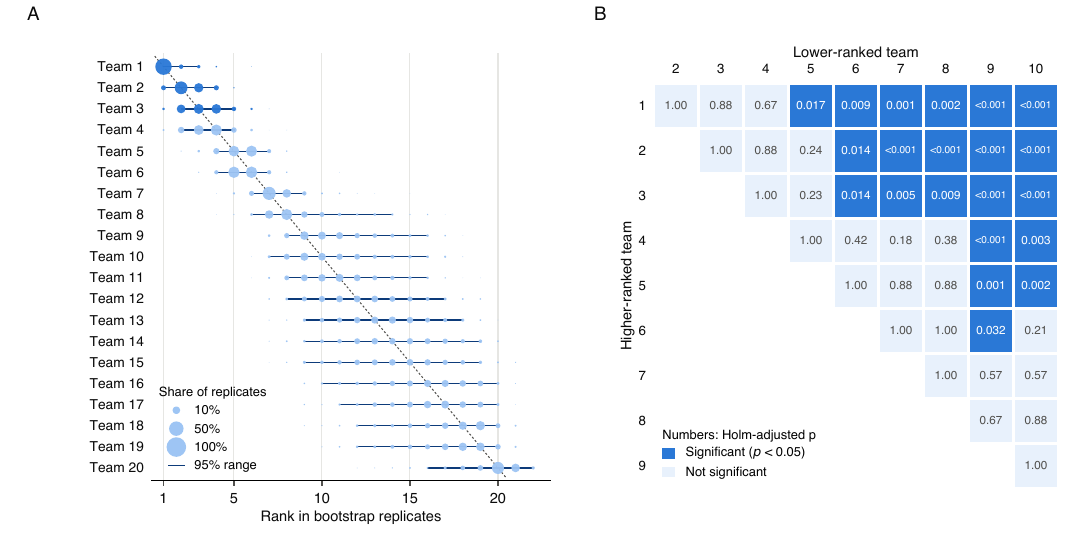}
\caption{Ranking stability, from 1,000 bootstrap replicates of the
300 test scans. A) Rank of the twenty best teams across replicates. Circle
area is the share of replicates with that rank, the line spans the central 95\%
of ranks, and the dashed diagonal marks the observed rank; the three winning
teams are shown in darker blue. B) Pairwise comparisons among the ten best
teams, labelled by rank. Each number is the Holm-adjusted $p$-value of a one-sided
paired Wilcoxon test that the higher-ranked team scored higher, and filled cells
are significant ($p < 0.05$). Confidence intervals of
the mean Dice are given in Table~\ref{tab:leaderboard}.}
\label{fig:ranking}
\end{figure*}

\section{Prompt sensitivity analysis examples}
\label{app:prompt-examples}
\setcounter{figure}{0}
Figure~\ref{fig:prompt-examples} shows three test findings from the prompt
sensitivity analysis (Section~\ref{sec:prompt-exp}), each segmented by Team 2
with its original sentence and with several edited versions of it. The rows
include each kind of location edit, a wrong size and a finding name without
location, and the Dice of each mask is given below it. Teams 2, 3
and 4 responded to the edits in the same way, and we show Team 2 because it
used no location post-processing, so its masks reflect how the model itself
responds to the text. Figure~\ref{fig:prompt-examples-t1} shows the same
findings for Team 1, which more often returned an empty mask or kept the mask on
the original finding.

\begin{figure*}[!tp]
\centering
\includegraphics[width=\textwidth]{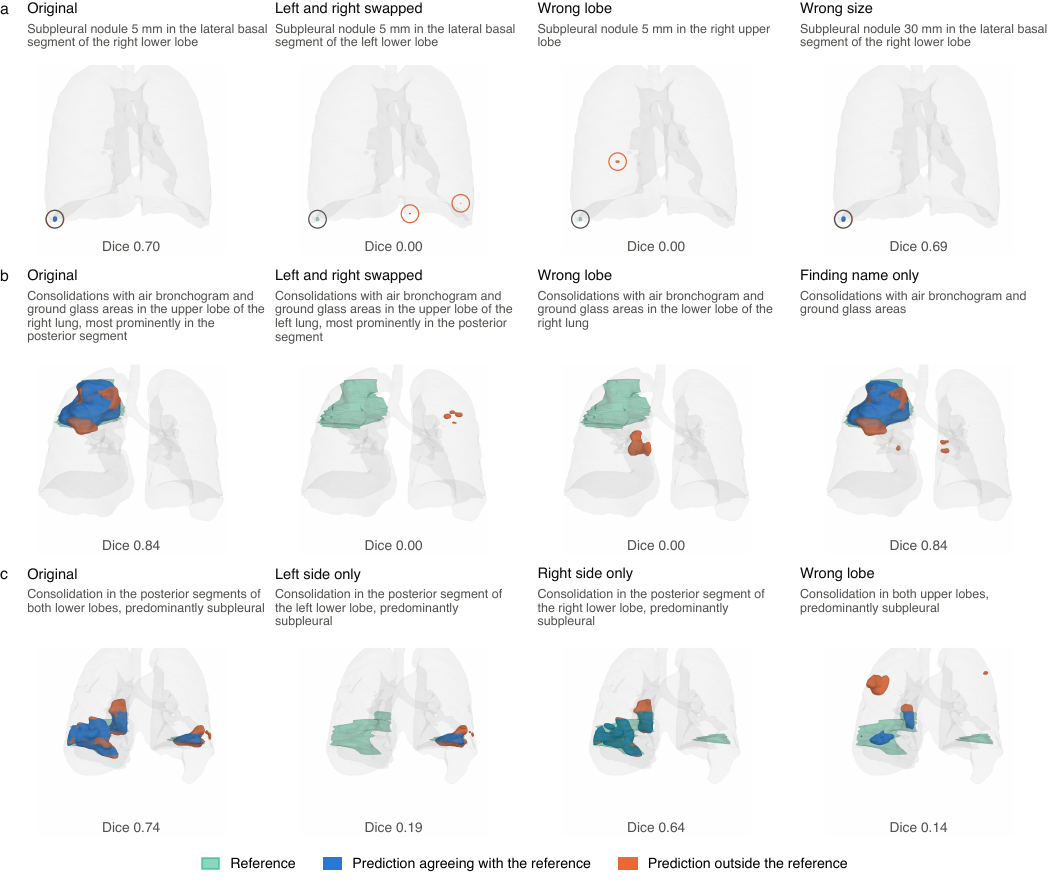}
\caption{Examples from the prompt sensitivity analysis (Team 2). Each row shows
one test finding segmented with its original sentence (left) and with edited
sentences. Masks are rendered in 3D inside the lungs and viewed from the front
(patient's right on the left). The reference is shown in translucent green, and
predicted surfaces are blue where they lie within 3\,mm of the reference
and orange elsewhere. Circles mark masks smaller than 1\,mL. Dice is
computed against the unchanged reference, so a prompt for one side of a
bilateral finding (c) can cover at most part of it.}
\label{fig:prompt-examples}
\end{figure*}

\begin{figure*}[!tp]
\centering
\includegraphics[width=\textwidth]{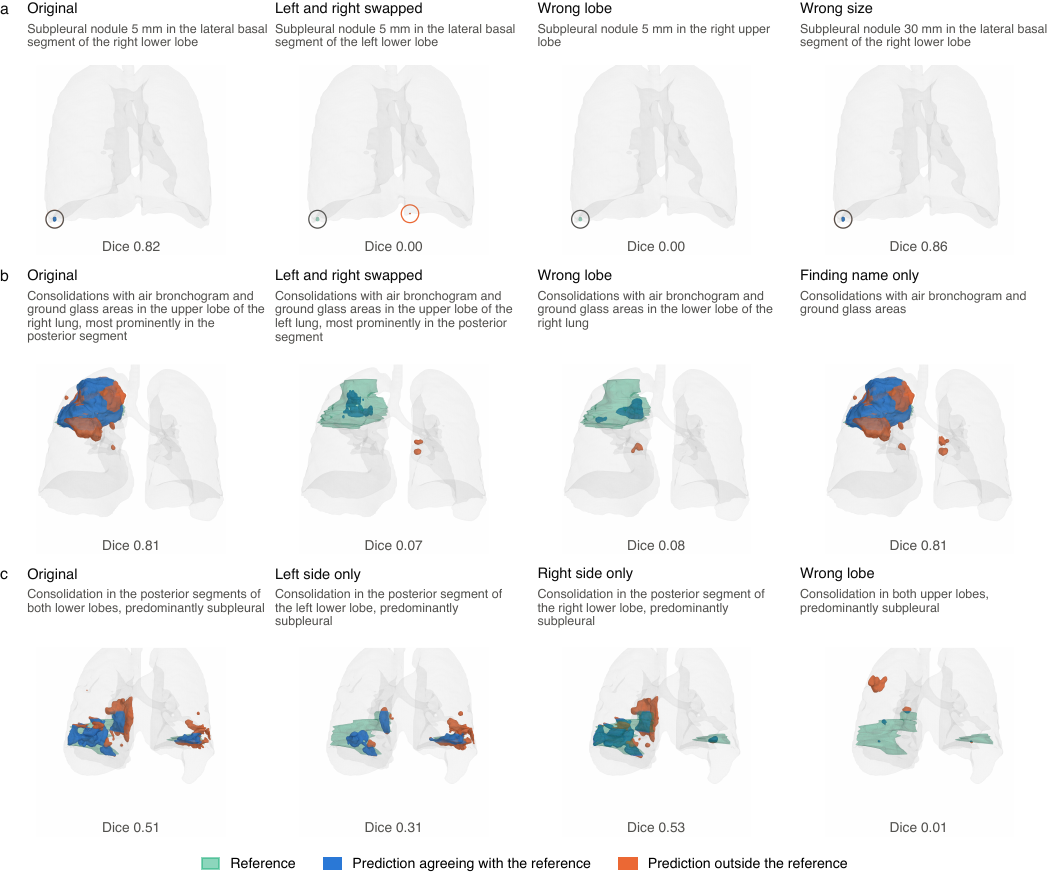}
\caption{The findings of Figure~\ref{fig:prompt-examples} segmented by Team 1.}
\label{fig:prompt-examples-t1}
\end{figure*}

\section{Performance by patient sex and age}
\label{app:sexage}
\setcounter{table}{0}\setcounter{figure}{0}
Table~\ref{tab:sexage} shows the performance of the ten best teams by patient
sex and age, using the per-scan results of their ranked submissions and the
sex and age recorded in CT-RATE. Dice was slightly higher for female than for
male patients in nine of the ten teams, but the difference was not significant
for any team (two-sided Mann--Whitney U test on per-scan Dice, $p \geq 0.16$).
Dice was highest in patients under 40 years and lower in older patients, a
difference that was also not significant for any team (Kruskal--Wallis test,
$p \geq 0.05$) and that partly reflects a higher share of non-focal findings in
patients aged 60 and over (16\% against 11--12\%). The tests were not corrected
for multiple comparisons.

\begin{table}[!htbp]
\centering
\caption{Performance of the ten best teams by patient sex and age group on the
300 test scans. Values are the mean over the ten teams, with the range across
teams in brackets.}
\label{tab:sexage}
\renewcommand{\baselinestretch}{1}\footnotesize
\setlength{\tabcolsep}{1.5pt}
\renewcommand{\arraystretch}{1.2}
\begin{tabular}{lcccc}
\toprule
Group & Scans & Findings & Dice & Hit rate \\
\midrule
Female & 127 & 243 & 0.359 {\scriptsize[0.330--0.384]} & 0.771 {\scriptsize[0.704--0.811]} \\
Male & 173 & 339 & 0.346 {\scriptsize[0.330--0.369]} & 0.732 {\scriptsize[0.678--0.767]} \\
\midrule
Under 40 years & 142 & 249 & 0.376 {\scriptsize[0.351--0.403]} & 0.783 {\scriptsize[0.727--0.811]} \\
40--59 years & 113 & 218 & 0.328 {\scriptsize[0.314--0.346]} & 0.736 {\scriptsize[0.670--0.771]} \\
60 years and over & 45 & 115 & 0.332 {\scriptsize[0.293--0.363]} & 0.697 {\scriptsize[0.643--0.730]} \\
\bottomrule
\end{tabular}
\end{table}

\section{BIAS checklist}
\label{app:bias}
\setcounter{table}{0}
Table~\ref{tab:bias} lists where this paper reports each item of the BIAS
reporting guideline \citep{maier2020bias}.

\begin{table*}[!p]
\centering
\caption{Location of each BIAS checklist item in this paper.}
\label{tab:bias}
\renewcommand{\baselinestretch}{1}\scriptsize
\setlength{\tabcolsep}{4pt}
\renewcommand{\arraystretch}{1.12}
\begin{tabularx}{\textwidth}{l>{\raggedright\arraybackslash}p{0.38\textwidth}>{\raggedright\arraybackslash}X}
\toprule
Item & Description & Location \\
\midrule
1 & Title conveys the essential information on the challenge & Title \\
2 & Abstract: purpose, design and main results & Abstract \\
3 & Keywords & Keywords \\
4a & Biomedical motivation & Section~\ref{sec:intro} \\
4b & Technical motivation & Section~\ref{sec:intro} \\
4c & Purpose of the challenge and gap addressed & Section~\ref{sec:intro} \\
5a--b & Challenge name and acronym & \ref{app:organization}, Name and organizers \\
6 & Organizing team & \ref{app:organization}, Name and organizers \\
7 & Life-cycle type & \ref{app:organization}, Life cycle and platform \\
8a--c & Event, platform and website & \ref{app:organization}, Life cycle and platform \\
9a & Allowed user interaction & Section~\ref{sec:organization} \\
9b & Policy on training data and pretrained models & Section~\ref{sec:organization} \\
9c & Participation of organizers & \ref{app:organization}, Participation, awards and publication \\
9d & Award policy & \ref{app:organization}, Participation, awards and publication \\
9e & Publication of results & \ref{app:organization}, Participation, awards and publication \\
10a--b & Submission method and frequency & \ref{app:organization}, Submission; Section~\ref{sec:assessment}, Ranking \\
11 & Timeline & \ref{app:organization}, Timeline \\
12 & Ethics approval & \ref{app:organization}, Ethics, data use, code \\
13 & Data usage agreement & \ref{app:organization}, Ethics, data use, code \\
14a--b & Code availability (organizers and participants) & \ref{app:organization}; Declarations; \ref{app:methods} \\
15 & Conflicts of interest & Declarations \\
16 & Author contributions & Declarations \\
17 & Field of application & Section~\ref{sec:mission}, Task \\
18 & Task category & Section~\ref{sec:mission}, Task \\
19a--b & Target cohort and challenge cohort & Section~\ref{sec:data}, Cohort \\
20 & Imaging modality & Section~\ref{sec:data}, Cohort \\
21 & Context information & Section~\ref{sec:mission}, Task \\
22a--b & Data origin and algorithm target & Section~\ref{sec:data}, Cohort \\
23 & Property to be optimized & Section~\ref{sec:mission}, Assessment aim \\
24a--d & Acquisition devices, protocols, centres and operators & Section~\ref{sec:data}, Data source and Cohort \\
25a & Definition of a case & Section~\ref{sec:data}, Data splits \\
25b & Number of training, validation and test cases & Section~\ref{sec:data}, Data splits \\
25c & Justification of the split sizes & Section~\ref{sec:data}, Data splits \\
25d & Characteristics of the splits & Section~\ref{sec:data}; Figure~\ref{fig:categories} \\
26a--d & Annotation method, instructions, annotators and merging & Section~\ref{sec:data}, Annotation \\
27 & Pre-processing & Section~\ref{sec:data}, Pre-processing \\
28a--b & Sources of error and inter-rater variability & Section~\ref{sec:data}, Sources of error \\
29a--b & Metrics and their justification & Section~\ref{sec:assessment}, Metrics \\
30a--c & Ranking method, missing results and justification & Section~\ref{sec:assessment}, Ranking \\
31a--b & Statistical methods and justification & Section~\ref{sec:assessment}, Statistical analysis \\
32 & Further analyses (methods) & Section~\ref{sec:further} \\
33a--b & Participating teams and method descriptions & Section~\ref{sec:participation}, Section~\ref{sec:top-methods}; \ref{app:methods} \\
34 & Metric values & Table~\ref{tab:leaderboard}; Table~\ref{tab:leaderboard-full} \\
35a & Rankings and number of test cases & Section~\ref{sec:ranking}; Table~\ref{tab:leaderboard} \\
35b & Results of the statistical analysis & Section~\ref{sec:stability}; Figure~\ref{fig:ranking} \\
36 & Results of further analyses & Section~\ref{sec:by-category}, Section~\ref{sec:qualitative}, Section~\ref{sec:further} \\
37 & Summary of the results & Section~\ref{sec:discussion} \\
38a--b & Biomedical and technical impact & Section~\ref{sec:discussion} \\
39a & Is the task solved? & Section~\ref{sec:discussion} \\
39b & Analysis of individual cases & Section~\ref{sec:qualitative}; Section~\ref{sec:discussion} \\
39c & Strengths and weaknesses of the methods & Section~\ref{sec:top-methods}; \ref{app:methods} \\
40 & Limitations & Section~\ref{sec:discussion} \\
41 & Future work & Section~\ref{sec:discussion} \\
\bottomrule
\end{tabularx}
\end{table*}

\end{document}